\def\year{2022}\relax
\documentclass[letterpaper]{article} 
\usepackage{aaai22}  
\usepackage{times}  
\usepackage{helvet}  
\usepackage{courier}  
\usepackage[hyphens]{url}  
\usepackage{graphicx} 
\usepackage{balance}
\usepackage{natbib}  
\usepackage{caption} 
\DeclareCaptionStyle{ruled}{labelfont=normalfont,labelsep=colon,strut=off} 
\usepackage{algorithm}
\usepackage{algorithmic}
\usepackage[utf8]{inputenc} 
\usepackage[T1]{fontenc}    
\usepackage{hyperref}       
\usepackage{url}            
\usepackage{booktabs}       
\usepackage{amsfonts}       
\usepackage{nicefrac}       
\usepackage{microtype}      
\usepackage{verbatim}       

\usepackage{newfloat}
\usepackage{listings}
\floatstyle{ruled}
\newfloat{listing}{tb}{lst}{}
\floatname{listing}{Listing}
\title{Set Twister for Single-hop Node Classification}
\author{
    Yangze Zhou, 
    Vinayak Rao, 
    Bruno Ribeiro
}
\affiliations{
  Purdue University\\
  West Lafayette, IN 47906 \\
  \{zhou950,varao\}@purdue.edu, ribeiro@cs.purdue.edu
}

\usepackage{bibentry}

\usepackage{amsthm}      
\usepackage{dsfont}
\usepackage{amsmath}
\usepackage{graphicx}
\usepackage[normalem]{ulem}
\usepackage{subcaption}
\usepackage{multirow}
\usepackage{natbib}
\usepackage{algorithm}
\usepackage{algorithmic}
\usepackage{thmtools}
\usepackage{thm-restate}

\usepackage{diagbox}

\usepackage{bm,xspace}

\newcommand{\Twister}{Set Twister\xspace}

\newcommand{\TwisterName}{Set Twister\xspace}

\newcommand{\DeepSets}{DeepSets\xspace}

\newtheorem{definition}{Definition}

\newtheorem*{proposition*}{Proposition}
\newtheorem{theorem}{Theorem}
\newtheorem*{theorem*}{Theorem}

\newtheorem*{corollary*}{Corollary}

\newtheorem{lemma}{Lemma}
\newtheorem*{lemma*}{Lemma}

\newenvironment{restat1}
{\restatable{theorem}{thmmain}}
{\endrestatable}

\newenvironment{restat2}
{\restatable{theorem}{thmexpressive}}
{\endrestatable}

\newenvironment{restatProp}
{\restatable{proposition}{univ}}
{\endrestatable}

\newenvironment{restatlemma}
{\restatable{lemma}{lembasis}}
{\endrestatable}

\newcommand{\cG}{\mathcal{G}}

\makeatletter
\let\overlinewithoriginalheight\overline
\newcommand*\overlinewithlessheight[1]{{\mathpalette\overline@aux{#1}}}
\newcommand*\overline@aux[2]{
  \begingroup
    \count0=\fam 
    \setbox0=\hbox{$\m@th #1\fam=\count0 #2$}
    \@tempdima=.4\ht0
    \setbox0=\hbox{$\m@th #1\fam=\count0\overlinewithoriginalheight{#2}$}%
    \advance\@tempdima by .6\ht0
    \ht0=\@tempdima 
    \usebox0
  \endgroup%
}
\let\overline\overlinewithlessheight

\let\underlinewithoriginaldepth\underline
\newcommand*\underlinewithlessdepth[1]{{\mathpalette\underline@aux{#1}}}
\newcommand*\underline@aux[2]{%
  \begingroup
    \count0=\fam
    \setbox0=\hbox{$\m@th #1\fam=\count0 #2$}%
    \@tempdima=.4\dp0%
    \setbox0=\hbox{$\m@th #1\fam=\count0\underlinewithoriginaldepth{#2}$}%
    \advance\@tempdima by .6\dp0%
    \dp0=\@tempdima
    \usebox0%
  \endgroup%
}
\let\underline\underlinewithlessdepth
\makeatother

\newcommand*\dbar[1]{{\overline{#1}}}
\newcommand{\harrow}[1]{\mathstrut\mkern2.5mu#1\mkern-11mu\raise1.6ex\hbox{$\scriptscriptstyle\rightharpoonup$}}

\def\1{\bm{1}}

\def\vh{{\bm{h}}}

\def\vx{{\bm{x}}}
\def\vy{{\bm{y}}}

\DeclareMathAlphabet{\mathsfit}{\encodingdefault}{\sfdefault}{m}{sl}
\SetMathAlphabet{\mathsfit}{bold}{\encodingdefault}{\sfdefault}{bx}{n}

\def\gF{{\mathcal{F}}}

\def\sN{{\mathbb{N}}}

\def\sR{{\mathbb{R}}}

\usepackage{cleveref}  

\usepackage{xcolor}

\begin{document}

\maketitle

\begin{abstract}
Node classification is a central task in relational learning, with the current state-of-the-art hinging on two key principles: (i) predictions are permutation-invariant to the ordering of a node's neighbors, and (ii) predictions are a function of the node's $r$-hop neighborhood topology and attributes, $r \geq 2$.
Both graph neural networks and collective inference methods (e.g., belief propagation) rely on information from up to $r$-hops away.
In this work, we study if the use of more powerful permutation-invariant functions can sometimes avoid the need for classifiers to collect information beyond $1$-hop.
Towards this, we introduce a new architecture, the \TwisterName, which generalizes DeepSets (\citeauthor{zaheer2017deep}), a simple and widely-used permutation-invariant representation. 
\TwisterName theoretically increases expressiveness of DeepSets, allowing it to capture higher-order dependencies, while keeping its simplicity and low computational cost.
Empirically, we see accuracy improvements of \TwisterName over DeepSets as well as a variety of graph neural networks and collective inference schemes in several tasks, while showcasing its implementation simplicity and computational efficiency.
\end{abstract}
\section{Introduction}

In this work, we consider learning permutation-invariant representations over finite but  variable-length sequences (e.g., a node's 1-hop   neighbors).
This is an important inductive bias in machine learning tasks with graph-structured data, where permutation-invariance ensures that the ordering of the rows and columns of an adjacency matrix does not affect the final node representation. Deep learning architectures that ensure permutation-invariance generalize simple sum- and max-pooling operations, and allow significantly improved performance in a number of node classification tasks in relational learning.
In practice, such tasks typically rely either on graph neural networks~\cite{Kipf2016,Hamilton2017,velickovic2018graph,luan2019break} or on {\em collective inference}~\cite{jensen2004collective,moore2017deep,sen2008collective} (and related methods~\cite{huang2020combining}), where the labels of all nodes in a network are simultaneously predicted using a weak learner which mostly considers information only from the $1$-hop neighborhood, which then is coupled with a collective learning procedure that allows the label of a node $r$-hops away to influence the predictions, compensating for the weak predictive performance of the weak learners.

In this paper, we study if the use of more powerful permutation-invariant functions can eliminate the computational burden of collective inference, and allow the cost of prediction of a node to depend on its neighborhood size, rather than the network size.

A simple but powerful framework to approximate permutation-invariant functions of a sequence  $\vh$ of neighborhood feature vectors
is \DeepSets~\citep{zaheer2017deep}. 
Here, each element of $\vh$ is independently fed into a feed-forward neural network $\phi$, after which these embeddings are aggregated using a sum-pooling operation to get an intermediate permutation-invariant representation of $\vh$. 
Finally, a nonlinear function $\rho$, instantiated as another neural network, is applied to get the final representation.  
\citet{zaheer2017deep} show that with suitable choices of $\phi$ and $\rho$, any continuous permutation-invariant function over the input sequences can be approximated by the form $\rho(\sum_{j=1}^{n_\vh}\phi(\vh_j))$,  where $n_\vh$ denotes the length of the sequence and $\vh_j$ is the $j$th element in $\vh$. 
Thus with universal approximators $\rho$ and $\phi$, \DeepSets's transform-sum-transform architecture itself is a universal approximator of smooth permutation-invariant functions.

The simple form of \DeepSets, along with its theoretical properties~\citep{zaheer2017deep, wagstaff2019limitations, xu2018how}, have made it very attractive to practitioners. 
However, the learnability of the model still remains a problem if approximating the true target function requires higher-order interactions between the elements of $\vh$. 
Examples are found when the true target function is the empirical variance or some form of minimax pairwise norm of a sequence of numbers, and complex neighborhood user feature relationship in social networks.
Under the transform-sum-transform structure of \DeepSets, second and higher-order interactions between elements of $\vh$ must be learnt by $\rho(\cdot)$ from the summed representation $\sum_{j=1}^{n_\vh}\phi(\vh_j)$. 
This can present a significant challenge to optimization routines and model architecture~\citep{wagstaff2019limitations}, resulting in poor empirical performance, as noted by \citet{murphy2019janossy} and confirmed by our experiments.
While there has been work attempting to address this challenge~\citep{moore2017deep,murphy2019janossy,  pmlr-v97-lee19d}, these come at significant computational expense, making them unattractive to practitioners.

\paragraph{Contributions.}
We propose the \Twister framework  that extends \DeepSets to learn richer intermediate representations, while still having comparable computational cost. 
Our architecture modifies the transform-sum-transform architecture of \DeepSets, introducing an intermediate `twist' operator that allows the representations fed into $\rho$ to capture interactions between elements of the input {sequence}.
We justify our approach by defining $k$-ary permutation-invariant interactions between elements of a sequence, and showing that our architecture is a truncation of a series expansion of such $k$-ary functions.
In our experiments, we demonstrate how our framework offers an effective trade-off between statistical and computational performance in several tasks compared to a number of baselines from the literature.
For node-classification tasks, we show how the flexibility of \Twister can allow simpler single-hop predictions, unlike graph neural networks and collective learning schemes.
\section{The \TwisterName Architecture}
\label{sec:nt}

We start by detailing the \TwisterName architecture.
We are interested in representations of permutation-invariant functions over finite but arbitrary length sequences $\vh \in \cup_{j=1}^\infty \sR^{j \times d_\text{in}}$.
For node-classification tasks, each element of the sequence is the feature vector of a neighboring node.
Write $n_\vh$ for the length of the sequence $\vh$, which we then write as
 $\vh = (\vh_1,\ldots,\vh_{n_\vh})$, with $\vh_i \in \sR^{d_\text{in}}$.
 
\begin{definition}[Permutation-invariant functions] 
For any sequence $\vh \in \cup_{j=1}^{\infty}\sR^{j \times d_\text{in}}$ and any permutation $\pi$ of the integers $(1,\dotsc,n_\vh)$, define $\vh_\pi$ as the reordering of the elements of $\vh$ according to $\pi$. A function $\dbar{f}$ is permutation-invariant if it satisfies $\dbar{f}(\vh) = \dbar{f}(\vh_\pi)$ for all $\vh$ and $\pi$.
\end{definition}

\DeepSets~\citep{zaheer2017deep} defines a class of permutation-invariant functions of $\vh$ as
\begin{equation}
    \text{DS}(\vh)=\rho_\text{DS}\Big(\sum_{j=1}^{n_\vh}\phi(\vh_j)\Big).
    \label{eq:deepsets}
\end{equation}
Here $\rho_\text{DS}$ and $\phi$ are learnable functions, typically feed-forward neural networks (MLPs).
The term $\sum_{j=1}^{n_\vh}\phi(\vh_j)$ forms representations of the elements of $\vh$ and adds them together, thereby forming a permutation-invariant representation of the sequence $\vh$.
This representation is then transformed by a second function $\rho_\text{DS}$, and
for powerful enough $\phi$ and $\rho_\text{DS}$, equation~\eqref{eq:deepsets} can be shown to be a universal approximator of continuous permutation-invariant functions~\citep{wagstaff2019limitations,zaheer2017deep}.
 Despite this, \citet{murphy2019janossy} shows that simply adding up representations of the elements of $\vh$ cannot capture dependencies between elements of the sequence, so that \DeepSets effectively offloads the task of learning these high-order dependencies to $\rho_\text{DS}(\cdot)$.
Recovering higher-order relationships from the summed representations of the individual components can be challenging, with  $\rho_\text{DS}(\cdot)$ often failing to capture these high-order dependencies~\citep{murphy2019janossy} given finite datasets.
Our proposed \TwisterName architecture recognizes this, and allows users to model and capture such interactions. Importantly, as we make precise later, \TwisterName does this without incurring a significantly heavier computational burden than \DeepSets.

\begin{figure*}[t!!!]
    \centering
    \includegraphics[width=.99\textwidth]{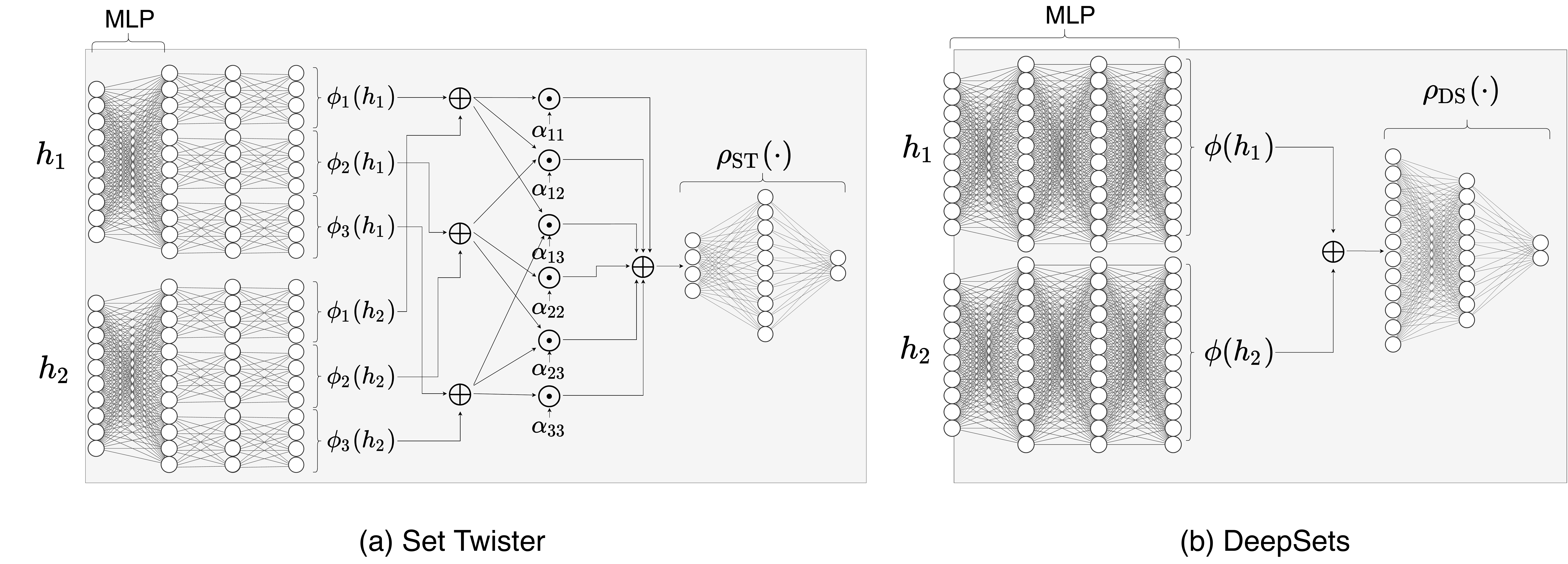}
    \caption{Illustration for input $\vh=(\vh_1,\vh_2) \in \sR^{2\times 10}$of (a) \Twister architecture (with $M=3$, $k=2$ and functions $\phi_i: \sR^{10} \to \sR^4$, $i=1,2,3$.) v.s.\ (b) {\DeepSets} architecture with same number of hidden layers and $\phi: \sR^{10} \to \sR^{12}$. The symbol $\oplus$ indicates the (element-wise) sum of the vectors.
    Note that the illustrated \Twister has fewer parameters than DeepSets.}
    \label{fig:illustration}
    \vspace{-10pt}
\end{figure*}

\begin{definition}[\TwisterName architecture] 
For integer-valued parameters $k$ and $M$ satisfying $1 \leq k \leq M$, the $k$-th order \TwisterName representation involves $M$ learnable representation functions  $\phi_1,\ldots,\phi_M$, with $\phi_i: \sR^{d_\text{in}} \to \sR^{d_\text{rep}}$, $d_\text{rep} \geq 1$, and takes the form
\begin{align}
 \text{ST}_{M,k}(\vh) &= \rho_\text{ST} \left(\dbar{f}_{M,k}(\vh) \right), \qquad \text{where}  \nonumber  \\
\dbar{f}_{M,k}(\vh)\! &= \! \sum_{u_1=1}^M \sum_{u_2=u_1}^M \!\!\!\cdots \!\!\! \sum_{u_k=u_{k-1}}^M \boldsymbol{\alpha}_{u_1,...,u_k}\odot\nonumber\\ &\left(\sum_{i=1}^{n_\vh}\phi_{u_1}(\vh_i)\right)\odot \cdots \odot \left(\sum_{i=1}^{n_\vh}\phi_{u_k}(\vh_i)\right).  \label{eq:kary}
\end{align}
\end{definition}
Here $\odot$ is the Hadamard product, $\rho_\text{ST}(\cdot)$ is a feed-forward network (MLP), and $\{\boldsymbol{\alpha}_{u_1,\ldots,u_k}\}_{u_1,\ldots,u_k}\in \sR^{d_\text{rep}}$ is a set of learnable parameter vectors.
\Cref{fig:illustration}(a) illustrates a \Twister architecture with $M=3$ and $k=2$, where $\phi_i$ has three hidden layers, $i=1,2,3$.
Observe from equation~\eqref{eq:deepsets} and \Cref{fig:illustration}(b) that DeepSets is just a special instance of \Twister with $M\!=\!k\!=\!1$.
In effect, \TwisterName produces $M$ intermediate representations, each in the form of DeepSets, but of smaller dimensionality (we recommend $1/M$th the dimensionality).
All size-$k$ subsets of these $M$ representations are then combined in a multi-linear fashion to produce the permutation-invariant representation $\dbar{f}_{M,k}(\vh)$.

\textbf{\Twister for node classification:}
Consider a graph $G=(\mathcal{V}, A,X)$, where $\mathcal{V}$ is the vertex set, $A$ is the adjacency matrix, and $X$ the node features. For any node $v\in \mathcal{V}$, write $\mathcal{N}(v)$ for the neighborhood set of node $v$ in $G$, then we can construct the sequence $\vh(v)$ as $\vh(v)=(X_u)_{u\in \mathcal{N}(v)}$. Since the order of the neighbors does not matter, we must learn a permutation-invariant function $\dbar{f}$ on $\vh(v)$, giving a permutation-invariant representation $\rho((\phi(X_v), \dbar{f}(\vh(v))))$ of  node $v$.
Specifically, a DeepSets representation for node $v$ will be
\begin{equation}
    \text{DS}^G(v)=\rho_\text{DS}\Big((\phi(X_v),\sum_{u\in \mathcal{N}(v)}\phi(X_u))\Big),
    \label{eq:deepsets-node}
\end{equation}
and a \Twister representation for node $v$ is
\begin{equation}
\begin{aligned}
         \text{ST}_{M,k}^{G}(v)=\rho_\text{ST}\Big(\phi_1(X_v),\cdots,\phi_M(X_v),\dbar{f}_{M,k}(\vh(v))\Big).
    \label{eq:twister-node}
\end{aligned}
\end{equation}

\textbf{Computational complexity:}
Let $d_\text{rep}$ be the dimensionality of the permutation-invariant representation produced by \Twister.
Then, \Twister's forward pass ---without $\rho_\text{ST}$--- requires
$O\left(Mn_\vh d_\text{in}d_\text{rep}+ {k+M-1\choose M-1} d_\text{rep}\right)$ 
operations.
The first term arises from computing the $M$ component representations $\phi_1(\vh_1),\phi_1(\vh_2),\dotsc, \phi_M(\vh_{n_\vh})$ of the elements in the input {sequence} $\vh$, while the second term arises from counting the number of terms in the summation in \Cref{eq:kary}.
We note that this scales linearly in the length of $\vh$: just like DeepSets, and unlike some other methods in the literature~\citep{murphy2019janossy}.
Importantly, in the context of node classificating, $n_\vh$ corresponds to the size of the neighborhood of a node, and not the size of the graph.
DeepSets (without $\rho_\text{DS}$), which corresponds to \Twister with $M\!=\!k\!=\!1$, requires 
$O(n_\vh d_\text{in}d_\text{rep} )$ operations.
In practice, to keep the number of parameters and operations comparable, a \Twister architecture (when $M\geq 2$) will involve lower-dimensional representations than a comparable DeepSets architecture. 
Writing the dimensionality of the DeepSets representation (i.e.\ the output dimension of $\phi$ in \Cref{eq:deepsets}) as $d_\text{DSrep}$,
we recommend setting $d_\text{rep} = d_\text{DSrep}/M$, using functions $\phi_i: \sR^{d_\text{in}} \to \sR^{d_\text{rep}}$.
With such a setting, the $\phi$ function of DeepSets can be viewed as a concatenation of the $M$ functions $\phi_i$ of \Twister.
Our architecture modifies DeepSets by breaking the output of $\phi$ into $M$ components, and following \Cref{eq:kary}, scrambles and recombines them.
The multilinear structure of this ``twisting'' operation is indexed by parameters $\alpha_{u_1,\dotsc,u_k}$.

In terms of number of parameters, the complexity for \Twister is
$O(Md_\text{in}d_\text{rep}+ {k+M-1\choose M-1} d_\text{rep})$.
In \Cref{fig:illustration}, \Twister ($M=3, k=2$) has $512$ operations ($572$ if we include $\rho_\text{ST}$) and $240$ learnable parameters counting all $\boldsymbol{\alpha}$ as free parameters ($300$ if we include $\rho_\text{ST}$), while \DeepSets has $828$ operations ($968$ if we include $\rho_\text{DS}$) and $408$ learnable parameters ($548$ if we include $\rho_\text{DS}$), nearly double the number of \Twister operations and parameters. 
In general, limiting \Twister to small values of $M$ and $k$ will keep its complexity close to, if not better than, \DeepSets.

\section{Analysis of \TwisterName}
\label{sec:gen_inst}

In what follows, we show that without  $\rho_\text{ST}(\cdot)$, \Twister is provably more expressive than \DeepSets.
The richer representation that \Twister learns results in reduced
reliance on $\rho_\text{ST}(\cdot)$, allowing it to be simpler than $\rho_\text{DS}(\cdot)$, and making \Twister easier to train: in most of our experiments, we find that it performs better
than \DeepSets.
\Cref{thm:main} presents the main theoretical result of our work,
justifying the multilinear structure of \TwisterName, and showing it allows us to approximate any $k$-ary permutation-invariant function.
\Cref{thm:k_expr} shows that the expressive power of this class of $k$-ary permutation-invariant functions increases with $k$, by virtue of being able to capture higher-order interactions among the elements of the input sequence. All theoretical proofs will be given in a longer version of the paper.
We start by defining this class of functions.
\begin{definition}[$k$-ary permutation-invariant functions] \label{def:kary}
Consider a sequence $\vh \in \cup_{j=1}^{\infty}\sR^{j \times d_\text{in}}$, 
and an integer $k \in \sN$. A $k$-ary permutation-invariant function $\dbar{f}^{(k)}$ takes the form
\begin{equation}
    \dbar{f}^{(k)}(\vh)=\sum_{i_1,i_2,...,i_k\in \{1,...,n_\vh\}}\vec{f}^{(k)}(\vh_{i_1},\vh_{i_2},...,\vh_{i_k}),
    \label{eq:equaapp}
\end{equation}
where $\vec{f}^{(k)}: \sR^{k \times d_\text{in}}\rightarrow \sR^{d_\text{rep}}$ is an arbitrary, possibly permutation-sensitive function.
The output of the function $\dbar{f}^{(k)}$ forms a $k$-ary representation of the input sequence $\vh$.
\end{definition}
Observe that $\dbar{f}^{(k)}$ above is insensitive to reorderings of elements of $\vh$, despite being composed of permutation-sensitive functions $\vec{f}$.
Further, note that \DeepSets produces a $k$-ary representation, with $k=1$. 
A closely related definition of $k$-ary permutation invariance was considered in~\citet{murphy2019janossy}, though in their implementation, computation scaled as $O(n_\vh^k)$,
quickly becoming impractical even for $k=2$. 
The next theorem shows that despite being linear in $n_\vh$, \Twister for any $k$ (as defined in \Cref{eq:kary}) can approximate any $k$-ary permutation-invariant function. 

\begin{restat1}
\label{thm:main}
Under mild assumptions, the limit $\lim_{M \to \infty} \dbar{f}_{M,k}(\cdot)$, with $\dbar{f}_{M,k}(\cdot)$ as in \Cref{eq:kary}, converges {\em in mean} to \Cref{eq:equaapp} for any $\dbar{f}^{(k)}(\cdot)$ if $\phi_1,\ldots,\phi_M$ are universal approximators (e.g., MLPs).
\end{restat1}

The next theorem shows the significance of \Cref{def:kary}, and why being able to approximate it is important:
briefly, the class of $k$-ary functions strictly increases with $k$.
We note once again that DeepSets, with $k=1$, produces $1$-ary representations, whereas $n_\vh$-ary representations can capture all possible permutation-invariant representations of a sequence.
\begin{restat2}
\label{thm:k_expr}
Assume $1< k \leq n_\vh$. Then, increasing $k$ in \Cref{eq:equaapp} 
strictly increases $\dbar{f}^{(k)}$'s expressive power, that is,
if  $\mathcal{F}_k$ is the set of all permutation-invariant functions of the form $\dbar{f}^{(k)}$, then $\mathcal{F}_{k-1}$ is a proper subset of $\mathcal{F}_k$. Thus, a $k$-ary permutation-invariant function $\dbar{f}^{(k)}$ can express any $(k\!-\!1)$-ary permutation-invariant function $\dbar{f}^{(k-1)}$, but the converse does not hold. 
\end{restat2}

The  proof of \Cref{thm:main} 
hinges on \Cref{lem:basis} below, which extends a result discussed in  \citet[Chapter $9$]{linearanalysis}. 

\begin{restatlemma}
\label{lem:basis} 
Define the $d_\text{rep}$-dimensional function $\vec{f}^{(k)}$ in \Cref{def:kary} as 
 ${\vec{f}^{(k)}(\vh_1,\ldots,\vh_k)} = \left(\vec{f}^{(k)}_1( \vh_1,\ldots,\vh_k),\dotsc,\vec{f}^{(k)}_{d_\text{rep}}( \vh_1,\ldots,\vh_k) \right)$. 
 Assume each component $\vec{f}_r^{(k)}$ is Riemann integrable on the domain $[a,b]^{k\times d_\text{in}}$ ($a,b\in \mathbb{R}$), and let $\{\gamma_u(\cdot)\}_{u=1}^\infty$ be orthogonal bases for Riemann integrable functions on $[a,b]^{d_\text{in}}$. Then, the set of the products
$   \{ \gamma_{u_1}(\cdot)\cdots\gamma_{u_k}(\cdot)$ , for $1\leq u_1\leq u_2 \leq \cdots \leq u_k \} $
is an {orthogonal} basis for any Riemann integrable function on $R=[a,b]^{k\times d_\text{in}}$.
The expansion coefficient of $r$-th component $\vec{f}^{(k)}_r(\cdot)$ is
\begin{equation}
\label{eq:alpha}
\begin{aligned}
        &\alpha^{(r)}_{u_1,\ldots,u_k}=c_{u_1,\ldots,u_k}\cdot\\&\frac{\int \cdots \int_R \vec{f}_r^{(k)}(\vh_1,\ldots,\vh_k)\gamma_{u_1}(\vh_1)\cdots\gamma_{u_k}(\vh_k)d\vh_1\cdots d\vh_k}{\int \cdots \int_R \gamma_{u_1}^2(\vh_1)\cdots\gamma_{u_k}^2(\vh_k) d\vh_1\cdots d\vh_k},
\end{aligned}
\end{equation}
where $c_{u_1,\ldots,u_k}\in \sR$ is a constant related to $\{u_1,\ldots,u_k\}$.
Thus the series expansion takes the form below (where convergence is in mean):
\begin{equation}
\label{eq:expand}
\begin{aligned}
    &\vec{f}_r^{(k)}(\vh_1,\ldots,\vh_k)\\&  = \lim_{M \rightarrow \infty} \sum_{u_1=1}^M \cdots  \sum_{u_k=u_{k-1}}^M \!\!\alpha_{u_1,...,u_k}^{(r)} \gamma_{u_1}(\vh_1) \cdots \gamma_{u_k}(\vh_k).
\end{aligned}
\end{equation}
\end{restatlemma}

Plugging the representation from \Cref{eq:expand} into \Cref{eq:equaapp},
and define $\boldsymbol{\alpha}_{u_1,...,u_k}=[\alpha^{(1)}_{u_1,...,u_k},\cdots,\alpha^{(d_\text{rep})}_{u_1,...,u_k}]^T$, $\boldsymbol{\vec{\gamma}}_{u_i}(\cdot)=[\gamma_{u_i}^{(1)}(\cdot),\cdots,\gamma_{u_i}^{(d_\text{rep})}(\cdot)]^T$, we can write out a series-expansion for the entire vector-valued function $\dbar{f}^{(k)}$:
\begin{equation}
\begin{aligned}
     \dbar{f}^{(k)}(\vh) &= \lim_{M\rightarrow \infty}\sum_{u_1=1}^M  \cdots  \sum_{u_k=u_{k-1}}^M \boldsymbol{\alpha}_{u_1,...,u_k}\\& \odot\left(\sum_{i=1}^{n_\vh}\boldsymbol{\vec{\gamma}}_{u_1}(\vh_i)\right)\odot \cdots \odot \left(\sum_{i=1}^{n_\vh}\boldsymbol{\vec{\gamma}}_{u_k}(\vh_i)\right).
\end{aligned}
\label{eq:hadamard}
\end{equation}
As before, the equations above converge in mean.

In \Twister, we truncate the above summation for some finite $M$.
Besides the obvious computational necessity of such a truncation, we justify this by three facts: 
1) in our overall architecture, the output of the truncated sum will be passed through another neural network $\rho_{ST}(\cdot)$ which can clean up the effects of the truncation approximation (see \Cref{prop:univ_apprx} below),
2) our neural network implementation seeks to find important bases that relate to the task, and doing this in a data-driven fashion can have smaller approximation error, and 
3) larger $M$ will lead to larger model capacity, corresponding to less smooth functions.
As far as the second point is concerned, it is possible to enforce orthogonality of the components $\gamma_i(\cdot)$, though we do not do this, since this only reduces the expressiveness of our finite truncation.
It is possible that enforcing orthogonality will lead to better identifiability and easier learning, however we leave investigating this for future work.

We formalize the first point below, showing that even with the finite truncation, the full \Twister architecture is a universal approximator of any continuous permutation-invariant function for any choices of $k$ and $M$.

\begin{restatProp} \label{prop:univ_apprx}
\Twister is a universal approximator of continuous permutation-invariant functions.
\end{restatProp}

\vspace{-.1in}
\section{Related Work}
\paragraph{Permutation invariance} Permutation invariance has been widely studied and applied in recent years, and is a basic requirement for applications involving set-valued inputs.
Examples include multiple instance learning~\citep{dietterich1997solving,maron1998framework}, the self-attention block (encoder) of the Transformer architecture used in NLP tasks~\citep{vaswani2017attention},
point cloud modeling~\citep{achlioptas2018learning,qi2017pointnet}
, and scene understanding and image segmentation~\citep{su2015multi,kalogerakis20173d,sridhar2019multiview}.

There are several different approaches to achieve permutation-invariance in the literature.
One line of work tries to find or learn a canonical orderings of input sequences related to the task at hand~\cite{niepert2016learning,Vinyals2016,zhang2019permoptim}, while difficulties of such approaches have been discussed in~\citet{murphy2019janossy}.
Another popular approach is data augmentation~\citep{cubuk2019autoaugment,fawzi2016adaptive}, where the training dataset is augmented with permutations of the training sequences, thereby biasing learning towards permutation invariance. 

Recent work~\citep{lyle2020benefits} argues that ``feature averaging'' is better than data augmentation, and justifies the most widely studied approach to permutation-invariance: by building permutation-invariance into the neural network architecture. 
As discussed earlier, DeepSets~\citep{zaheer2017deep} uses sum-pooling over instance-based embeddings and studies its theoretical properties, while \cite{qi2017pointnet,ravanbakhsh2016,edwards2017towards} propose similar approaches using other pooling operations such as max-pooling. 
\citet{wagstaff2019limitations} studies the theoretical limits of such approaches. \citet{murphy2019janossy} extends this framework to have functions with higher-order interactions. 
Set Transformer~\citep{pmlr-v97-lee19d} uses self-attention blocks of a Transformer architecture to model the interaction among elements in the input.
More broadly, $\cG$-invariant neural networks dealing with invariance according to general group actions are discussed in~\citep{cohen2016group,ravanbakhsh2017equivariance,kondor2018generalization,bloem2019probabilistic,maron2018invariant, maron2019universality}.

\paragraph{Graph neural networks and collective inference for node classification} Graph Neural Networks (GNNs) constitute a popular class of methods for learning vertex representations~\citep{abu2019mixhop,velickovic2018graph,xu2018,xu2018how,pmlr-v97-murphy19a} using neural network to capture information up to $r-$hop neighborhood topology. 
The aggregation layer of GNNs are permutation-invariant. Truncated Krylov GCN~\cite{luan2019break} leverages multi-scale information and improves the expressive power of GCNs~\cite{Kipf2016}. GRAND~\cite{feng2020grand} uses random propagation strategy to mitigate the issue of over-smoothing and non-robustness. However, the success of GNN is yet to be studied.

Collective inference~\cite{jensen2004collective,moore2017deep,sen2008collective} is popular in relational learning which incorporates dependencies among node labels and propagate inferences during training to strengthen poorly-expressive relational node classifiers. \citet{pmlr-v139-hang21a} propose a general procedure {\em collective learning} on GNNs to improve their expressive power. A related work \citet{huang2020combining} combines shallow models that ignore the graph structure with simple post-processing procedures to exploit correlation in the label structure, and exceed or match state-of-the-art GNNs. In the paper, we also want to see if more expressive $1$-hop representation is enough for existing node classification tasks.

\paragraph{Series decomposition} Our approach of truncating an infinite series representation follows a long tradition in statistics, engineering and machine learning, starting from Principle Component Analysis (PCA)~\citep{hotelling1933analysis}, through Fourier series decomposition  to more general orthogonal polynomials~\citep{math}. Popular applications of such ideas for function decomposition include kernel methods and Gaussian processes (e.g.\ \citet{williams2001nyst}). \citet{Kipf2016} use Chebyshev polynomial expansions to approximate graph convolutional networks in the context of of classifying nodes of graph.

\section{Experimental results}
\label{exp}
We empirically evaluate the performance of \Twister on a variety of tasks, in terms of prediction accuracy, mean absolute error (MAE) and computational complexity. 
We focus on $M=k=2$ for \Twister, which is one level of expressiveness above DeepSets, and has comparable compute-time to DeepSets. We first apply \Twister on a variety of synthetic datasets to empirically show our theoretical claims. Then, we apply \Twister on node classification tasks on well-known graph datasets, and compare with multiple graph neural networks and collective inference methods.

\begin{table*}[t!!!]
\centering
\caption{Test MAE in the variance task and test accuracy in the range \& maxmin task, using {image inputs}. Per-epoch training times are included. MC indicates a Monte Carlo approximation. Standard deviations computed over 5 random initializations are shown in parentheses. Best results (with statistical significance determined by a two-sample $t$-test at $p=0.05$) are in bold.
}
\label{table:extra}
\resizebox{0.79\textwidth}{!}{
\begin{tabular}{llrrrrrrr}
& &
\multicolumn{2}{c}{variance} & \multicolumn{2}{c}{range} & \multicolumn{2}{c}{maxmin} \\
\cmidrule(lr){3-4}
\cmidrule(lr){5-6}
\cmidrule(lr){7-8}
\multicolumn{1}{c}{Cost} & 
\multicolumn{1}{c}{Model} & 
\multicolumn{1}{c}{MAE $\downarrow$} &
\multicolumn{1}{c}{time (s/epoch)} &
\multicolumn{1}{c}{acc.$\uparrow$} &
\multicolumn{1}{c}{time (s/epoch)} &
\multicolumn{1}{c}{acc.$\uparrow$} &
\multicolumn{1}{c}{time (s/epoch)} \\
\toprule
\multirow{2}{*}{$O(n_\vh)$} & {\bf \Twister ($M\!=\!\!k\!=\!2$)}
  & 0.200(0.007) & 2.386(0.020) & 0.931(0.002) & 2.441(0.036) & {\bf 0.931(0.003)} & 2.428(0.049) \\
& DeepSets  & 0.343(0.012) & 1.804(0.015) & {\bf 0.940(0.001)} & 1.842(0.012) &  0.912(0.003) & 1.839(0.012)   \\
\hline
\multirow{2}{*}{$O(n_\vh^2)$} & JP 2-ary  & 0.803(0.016) & 1.865(0.011) & 0.885(0.004) & 1.836(0.019)& 0.752(0.008) &  1.852(0.011)\\
 &Set Transformer & 0.327(0.020) & 6.395(0.037) & 0.917(0.002) & 6.526(0.036)  & 0.825(0.010) & 6.607(0.029)  \\
\hline
MC & JP Full (GRU)  &  {\bf 0.176(0.006)} & 3.993(0.021) & 0.932(0.002) & 4.077(0.087) & 0.838(0.020) & 4.085(0.041) \\
\bottomrule
\end{tabular}
}
\vspace{-5pt}
\end{table*}

\subsection{Synthetic tasks to validate theoretical claims}
In this subsection, we consider  tasks for which we know the task's high-order dependencies. We compare \Twister's performance against other widely used permutation-invariant methods: \DeepSets~\citep{zaheer2017deep}, $2$-ary Janossy Pooling~\citep{murphy2019janossy}, Full Janossy Pooling using GRUs with attention mechanisms (JP Full)~\citep{meng2019hats} and Set Transformer without inducing points~\citep{pmlr-v97-lee19d}.
We consider Full JP to be the current state-of-the-art if the required compute time and the randomness of its representations are not practical issues for the task.

\subsubsection{Arithmetic tasks: End-to-end learning over images}
We consider three permutation-invariant arithmetic tasks on integer sequences: Computing the empirical {\em variance} and {\em range} (from \citet{murphy2019janossy}), and a new task, the {\em maxmin} task. 
The inputs are sequences of $28\times28$ images from InfiMNIST \citep{loosli2007training}, an ``infinite'' dataset
containing images of digits 0-9, obtained using pseudo-random deformations and translations of MNIST. 

Consider a sequence $\vh$ consisting of $n_\vh$ images of digits from the InfiMNIST dataset. 
Denote by $\vy_i$ the label of the $i$-th image $\vh_i$ in the sequence.
The {\em variance} task receives a sequence $\vh$ of $5$ images drawn uniformly with replacement from the dataset and predicts the variance of the image labels $\frac{1}{2n^2_\vh}\sum_{i,j}(\vy_i-\vy_j)^2$; the {\em range} task receives a sequence $\vh$ of $5$ images distributed the same way and tries to predict their range (the difference between the maximum and minimum labels).
The {\em maxmin} task accepts a sequence of the same distribution and predicts $\max_i \min_j |\vy_i-\vy_j|$. 
The {\em variance} task involves learning $2$-ary dependencies, while the  {\em range} and {\em maxmin} tasks depend on the entire sequence (i.e., they are $n_\vh$-ary tasks). 
Both {\em variance} and {\em range} tasks have linear time $O(n_\vh)$ solutions, while computing {\em maxmin} scales quadratically with the sequence length.

We randomly generate 100,000 training sequences, 10,000 validation sequences and 10,000 test sequences from the InfiMNIST data for each run.
We train our models in an end-to-end manner, treating the images as 784-dimensional vector inputs. 
We summarize the model architecture here. $\phi$ is set to be a MLP with 3 hidden layers, and $\rho$ is set to be a MLP with 1 hidden layer and a scalar output. The Janossy Pooling model implementation follows \citet{meng2019hats}. For Set Transformer, the hidden dimension is set to be $128$, and number of attention heads is $4$.
We optimize all the models using Adam with minibatch size $128$ and learning rate $1\times 10^{-4}$ for 2,000 epochs, with the exception of Set Transformer and Full Janossy Pooling, where a learning rate of $5\times 10^{-4}$ gives better results. 

For the integer-valued {\em range} and {\em maxmin} tasks, following \citet{murphy2019janossy}, we round the model outputs and report the test accuracy (i.e.\ the fraction of test data points whose output is correctly predicted).
For the {\em variance} task, we report the test MAE. 
We also include running times per epoch in \Cref{table:extra}. 
To make the comparison clearer and fairer, we divide algorithms in terms of their computational cost, with MC denoting a Monte Carlo approximation.

From \Cref{table:extra}, we can see that \Twister is among the best methods in the {\em range} and {\em maxmin} tasks, and in the {\em variance} task is second only to Full Janossy Pooling (and even in the last case, there is a significant gap between these two methods and all others).
DeepSets obtained the highest accuracy in the {\em range} task, but struggled considerably for the variance test, where the 2-ary structure of \Twister allowed it to easily learn the task.
We emphasize that though we consider \Twister with $k=2$, our method performs well on the $n_\vh$-ary  {\em range} and {\em maxmin} tasks.
This is due to the nonlinear neural network $\rho$ at the output (see also \Cref{prop:univ_apprx}).
\DeepSets also includes such a neural network, however other than the {\em range} task, its performance is worse.
This demonstrates our point that despite being a universal approximator, DeepSets can struggle with practical learning tasks.
By incorporating higher-order representations, \Twister requires a simpler $\rho(\cdot)$, and results in easier learnability, and significantly improved performance. In terms of runtime, \Twister is about twice as fast as Set Transformer (the most expensive model) and JP Full, and comparable to \DeepSets.

\begin{table*}[h!!!]
\centering
\caption{Test accuracy and per-epoch training times in range and maxmin tasks and test MAE in the variance task over randomly encoded inputs. MC indicates a Monte Carlo approximation. Standard deviations computed over 5 random initialization runs are shown in parentheses. Best results (with statistical significance determined by a two-sample $t$-test at $p=0.05$) are in bold. }
\label{table:1}
\resizebox{0.79\textwidth}{!}{
\begin{tabular}{llrrrrrrrr}
&&
\multicolumn{2}{c}{variance} & \multicolumn{2}{c}{range} & \multicolumn{2}{c}{maxmin} \\
\cmidrule(lr){3-4}
\cmidrule(lr){5-6}
\cmidrule(lr){7-8}
\multicolumn{1}{c}{Cost} & 
\multicolumn{1}{c}{Model} & 
\multicolumn{1}{c}{MAE $\downarrow$} &
\multicolumn{1}{c}{time (s/epoch)} &
\multicolumn{1}{c}{acc.$\uparrow$} &
\multicolumn{1}{c}{time (s/epoch)} &
\multicolumn{1}{c}{acc.$\uparrow$} &
\multicolumn{1}{c}{time (s/epoch)} \\
\toprule
\multirow{2}{*}{$O(n_\vh)$} &
  {\bf \Twister ($M\!=\!\!k\!=\!2$)}
  & {\bf 0.333(0.008)} & 2.072(0.011) & 0.944(0.004) & 2.105(0.018) & 0.705(0.011) & 1.746(0.017)\\
   & DeepSets & 0.417(0.015) & 1.501(0.016) &  0.943(0.004) & 1.504(0.006) & 0.615(0.009) & 1.246(0.003)\\
  \hline
\multirow{2}{*}{$O(n_\vh^2)$} & 
 JP 2-ary  & 1.276(0.079) & 1.566(0.018) & 0.936(0.016) & 1.525(0.016) & 0.612(0.015)& 1.248(0.006)\\
 & Set Transformer  & {\bf 0.316(0.020)} & 6.461(0.031) & {\bf 0.999(0.001)} & 6.576(0.02) &  0.587(0.017) & 6.636(0.007)\\
 \hline MC &
  JP Full (GRU) & {\bf 0.355(0.041)} & 4.040(0.055) & {\bf 0.999(0.000)} & 3.742(0.072) & {\bf 0.784(0.031)}& 4.356(0.032)\\
\bottomrule
\end{tabular}
}
\vspace{-1pt}
\end{table*}

\subsubsection{Arithmetic tasks over larger vocabulary sizes}

Our next experiment is more challenging, increasing the vocabulary size for our previous tasks from $0-9$ to $0-99$. 
For the {\em variance} and {\em maxmin} tasks, we input a sequence $\vx$ of $10$ integers drawn uniformly with replacement from $\{0,1,...,99\}$, and for the {\em range} task, a sequence of $5$ integers distributed the same way. 
In all tasks, rather than working with images of the digits, we now assign each number of random $100$-dimensional embedding.  
As before, we randomly generate 100,000 training sequences, 10,000 validation sequences and 10,000 test sequences for each run.

The model architecture is similar to the previous setting thus omitted. We optimize all models using Adam with learning rate $5\!\times\! 10^{-4}$ for $2,000$ epochs and minibatch size 128 for the {\em range} and {\em variance} tasks. 
For the {\em maxmin} task, we run all the models except Set Transformer and Full Janossy Pooling for $40,000$ epochs using SGD with momentum, while Adam is used for Set Transformer and Full Janossy Pooling.
We also use a learning rate scheduler. 
As before, we report test accuracy of the {\em range} task and test MAE of the {\em variance} task in \Cref{table:1}, along with run times per epoch in \Cref{table:1}.

For the {\em variance} task involving 2-ary dependencies, \Twister, Set Transformer, and JP full perform best. 
DeepSets performs significantly worse in this case, struggling to capture the 2-ary interactions between the elements. 
Importantly, \Twister is only slightly slower than DeepSets, and considerably faster than JP Full and Set Transformer.
While \Twister can be viewed as an approximation to JP 2-ary, it outperforms it significantly, in part due to difficulty training that model. 
For the $n_\vh$-ary {\em range} task, all models perform relatively well. 

For the {\em maxmin} task, JP Full performs best, followed by \Twister. All other models lag significantly behind these two, with Set Transformer performing worst.
JP Full's performance now is in contrast to its poor performance on the same task with the real image data, in part due to the availability of longer sequences: \Twister's performance is relatively insensitive to this setting. 
In additional results that are not presented, \Twister is also relatively robust to settings of $M$, both in terms of predictive accuracy as well as run time, which means in the simplest case $M=2$ is enough to capture the dependencies in these tasks.

\begin{table*}[h!!!]
\centering
\caption{Test accuracy (in \%) for set representation model comparing with other GNN and collective inference models on six relational data. Standard deviations computed over 5 random initializations are shown in parentheses. Best results are in bold.}
\label{tabel:graph-res}
\resizebox{0.8\textwidth}{!}{
\begin{tabular}{lcccccccccc}
Model & Friendster & PubMed & Cora & Citeseer & Arxiv & wikiCS\\  
\hline
GCN & 34.19(00.49) & 86.82(00.27) & 88.39(00.95) &  76.78(01.73) & 54.15(00.43) & 76.53(00.71) \\ 
GAT & 32.28(00.74) & 86.54(00.36) & 87.95(00.55) & 76.75(00.72) & 54.65(00.32) & 77.65(00.11)\\
TK-GCN & 33.79(00.38) & 86.01(00.18) & 87.76(00.82) & 76.09(00.61)& 63.47(01.24) & 78.77(00.50)\\
TK-GCN+CL & 32.75(00.68) & 86.12(00.14) & 88.13(00.36) &  77.23(00.63) & 63.27(01.28) & 78.83(00.43)\\
GRAND & 34.48(00.94) & 86.29(00.57) & 87.95(00.75) & 75.97(01.18) & 63.39(00.84) & 77.03(00.59)\\
\hline
DeepSets & 34.48(00.94) & 89.85(00.37) & 87.06(00.68) & 74.20(01.76) & 66.12(00.13) & 77.90(00.59)\\
Set Twister ($M\!=\!\!k\!=\!2$) & {\bf 37.34(01.17)} & {\bf 90.15(00.29)} & 82.74(00.34) &  71.43(02.09) & 64.75(00.13) & 76.58(00.77)\\
\hline
C\&S~\cite{huang2020combining}  & --- & 89.74\phantom{(00.00)} & {\bf 89.05}\phantom{(00.00)} & 76.22\phantom{(00.00)} & 70.60\phantom{(00.00)} & 82.54\phantom{(00.00)}\\
\hline
SOTA & --- & {\bf 90.30}\phantom{(00.00)} & 88.49\phantom{(00.00)} & {\bf 77.99}\phantom{(00.00)} & {\bf 73.65}\phantom{(00.00)} & {\bf 82.56}\phantom{(00.00)}\\
\bottomrule
\end{tabular}
}
\vspace{-5pt}
\end{table*}

\subsection{Node classification tasks}

In this section, we apply our methods on six graph datasets (\Cref{tabel:stat}) for single-hop node classification using permutation-invariant representation, and compare against state-of-the-art (SOTA) graph neural networks and collective inference methods. The Friendster dataset is a social network graph~\cite{teixeira2019GNNmiscalibrated}; the Pubmed, Cora and Citeseer are three classic citation network benchmarks~\cite{namata2012query}; the Arxiv is from the Open Graph Benchmark (OGB)~\cite{hu2020ogb}; and wikiCS is a web graph~\cite{mernyei2020wiki}.

\begin{table}[t!!]
\vspace{-5pt}
\centering
\caption{Summary statistics of graph datasets.}
\label{tabel:stat}
\resizebox{0.3\textwidth}{!}{
\begin{tabular}{lrrr}
Datasets & Classes & Nodes & Edges \\  
\hline
Friendster & 4 & 43,880 & 145,407 \\
Pubmed & 3 & 19,717 & 44,338\\
Cora & 7 & 2,708 & 5,429\\
Citeseer & 6 & 3,327 & 4,732\\
Arxiv & 40 & 169,343 & 1,166,243\\
wikiCS & 10 & 11,701 & 216,123\\
\bottomrule
\end{tabular}
}
\vspace{-10pt}
\end{table}

\paragraph{Data splits.} The training/validation/test split for Friendster dataset follows \citet{teixeira2019GNNmiscalibrated}, and all the splits for other datasets are the same as \citet{huang2020combining}. Specifically, for wikiCS, there are $20$ different training/validation splits and the test performances are averaged over these splits.

\paragraph{Comparison models.} For permutation-invariant representations, using the input features as the raw node features, we apply DeepSets and \Twister (reporting best results with {\em sum} or {\em mean} aggregation). For each node, we consider the neighborhood features as a sequence, obtain the permutation-invariant representation of the sequence and concatenate with the node feature itself to get node representations (as shown in \Cref{eq:deepsets-node,eq:twister-node}). We do not implement Janossy Pooling due to memory issues. Set Transformer is not designed for variable-size sequence, and we use an attention-based GNN graph attention network~\cite{velickovic2018graph} (GAT) for comparison. Other GNN models include graph convolutional network~\cite{Kipf2016} (GCN), truncated Krylov GCN~\cite{luan2019break} (TK-GCN), and GRAND~\cite{feng2020grand} which achieve good performance on several semi-supervised node classification tasks. We also compare with a recent collective inference method~\cite{pmlr-v139-hang21a}, which can be applied on any GNN methods to improve the expressive power, here we use TK-GCN as a base GNN model (TK-GCN+CL). Since we have the same data splits as \citet{huang2020combining}, we present the best reported accuracy from their C\&S model, and SOTA results from other models reported in \citet{huang2020combining} as baselines.

\paragraph{Implementation details.} The number of hidden layers was chosen between \{$2,3$\}. The number of neurons in the hidden layers was chosen between \{$16,32$\} for GAT (with $8$ attention heads) and \{$128,256,512$\} for all other models. For all models we used Dropout with probability $0.5$. We optimized all models using Adam with learning rate chosen from \{$1\!\times\! 10^{-3},5\!\times\! 10^{-3},1\!\times\! 10^{-2}$\} and strength of weight decay was set as $5\!\times\! 10^{-4}$. We trained all models in a full-batch end-to-end manner except C\&S and SOTA reported from \citet{huang2020combining}. Our results show the test accuracy from the model achieving best validation accuracy. 

\paragraph{Results.} The test accuracy results are reported in \Cref{tabel:graph-res}. Note that the last two rows are reported results from \citet{huang2020combining}, thus do not have standard deviations. All the other models are trained in an end-to-end manner. Because of the highly imbalanced label distribution in the Friendster dataset, we reweight the test accuracy by the label proportion as in \citet{teixeira2019GNNmiscalibrated}. We see that for the  Friendster and PubMed datasets, single-hop node classification using Set Twister achieves the best results, while in Cora and Citeseer, \Twister performs the worst. One reason for this is that as seen in \Cref{tabel:stat}, the Cora and Citeseer graphs have small degrees, and, thus, there are almost no higher-order dependencies to capture in the node neighborhoods. In the Arxiv dataset, which is large and dense, with similar architectures, DeepSets and Set Twister outperform most GNN and collective inference methods. 
We note that for the GCN and Arxiv dataset, we obtain performance results that are worse than those reported in \citet{hu2020ogb}, and choose instead to present our results.

A takeaway is that in graphs with reasonably large neighborhoods, by only using $1$-hop neighborhood information rather than the whole graph structure ($r$-hop information, for $r \geq 2$) and a powerful permutation-invariant classifier, we can achieve node classification performance comparable to widely-used GNN classifiers in most tasks, and to state-of-the-art methods in some of the tasks.

\section{Conclusion}
This work introduced the \Twister representation architecture and evaluated its performance in node classification tasks, where we show set representations of immediate neighbors ($1$-hop information) perform as well as widely-used GNN methods, and sometimes on par with state-of-the-art methods.
We argue that this is due to the increased representational power that \Twister affords:
\Twister extends the aggregated instance-based representation used in DeepSets~\citep{zaheer2017deep} to capture higher-order dependencies in the input sequence.
\Twister's architecture is theoretically justified by connecting it to series approximations of permutation-invariant functions.
In \Twister's simplest form $(M=2,k=2)$, our empirical results showed significant accuracy improvements over DeepSets in most synthetic tasks, while being nearly as fast.
\Twister was also contrasted with slower, more complex methods such as Janossy Pooling $k=2$ and $k=n_\vh$ (full) and Set Transformer, where the latter two are nearly 300\% to 600\% slower per epoch than DeepSets, respectively.
Here, \Twister $(M=2,k=2)$ tends to obtain comparable results (being the best method in some tasks, and among the best in others).
In the near future, we expect extensions of \Twister whose algorithmic innovations will allow practitioners to explore larger $k$ and $M$ values, while still maintaining computation and memory efficiency as $M=2,k=2$, especially in applications involving longer input sequences.

\balance
\bibliography{approx,rp_bibs,ryan_bib,ribeiro,node_class}

\newpage
\section*{Supplementary Material of ``Set Twister for Single-hop Node Classification''}

\appendix
\section{Proof of results}

In order to prove \Cref{thm:main}, we first introduce \Cref{thm:2ary}, which is discussed in \citet[Chapter $9$]{linearanalysis}. 

\begin{theorem}[\citet{linearanalysis}]
\label{thm:2ary}
If $\vec{f}(x,y): \mathbb{R}^2\rightarrow \mathbb{R}$ is a Riemann integrable function on a rectangle $[a,b]\times[c,d]$ ($a,b,c,d\in \mathbb{R}$), and $\{f_i(x)\}$ and $\{g_j(y)\}$ are orthogonal bases for the Riemann integrable functions on $[a,b]$ and $[c,d]$, then the set of the products
\begin{equation}
    \{f_i(x)g_j(y)\},\ i=1,2,...,j=1,2,...
\end{equation}
is a basis for any Riemann integrable function on $R$, where $R$ is the rectangle $a\leq x\leq b , c\leq y\leq d$. 

The generalized Fourier coefficients of any Riemann integrable function $\vec{f}$ on R is
\begin{equation}
    \alpha_{i,j}=\frac{\iint_R \vec{f}(x,y)f_i(x)g_j(y)dR}{\iint_R f_i^2(x)g_j^2(y) dR}.
\end{equation}
Thus the series expansion of $\vec{f}(x,y)$ can be written as 
\begin{equation}
    \sum_{i,j=1}^\infty \alpha_{i,j}f_i(x)g_j(y)
\end{equation}
This series converges {\em in mean} to $\vec{f}(x,y)$.
\end{theorem}

The proof for this theorem can be found in \citet[Chapter $9$]{linearanalysis}. 
As pointed out in \citet{linearanalysis}, it can be applied for functions of any finite number of variables. 
A following \Cref{lem:basis} is introduced in our paper, which leads to the proof of \Cref{thm:main}. 
Note that we did not include the following constants $c_{u_1,\ldots,u_k}\in \sR$ in \Cref{eq:alpha} of \Cref{lem:basis} in the main manuscript, but make them explicit below.

\lembasis*

\begin{proof}
\Cref{thm:2ary} can be extended to functions of any finite number of variables. So if $\{f_i(x)\}$ are orthogonal bases for the Riemann integrable functions on $[a,b]$, then the set of the products
\begin{equation}
    \{f_{i_1}(x_1)f_{i_2}(x_2)\cdots f_{i_{kd_\text{in}}}(x_{kd_\text{in}})\},\ i_1,i_2,...,i_{kd_\text{in}}=1,2,...
\end{equation}
is an orthogonal basis for any Riemann integrable function on $[a,b]^{k\times d_\text{in}}$. 

Also $\{f_{i_1}(x_1)f_{i_2}(x_2)\cdots f_{i_{d_\text{in}}}(x_{d_\text{in}})\},\ i_1,i_2,...,i_{d_\text{in}}=1,2,...$ is an orthogonal basis for any Riemann integrable function on $[a,b]^{d_\text{in}}$, define it as $\gamma_u(\vh), \ u=1,2,...$, where $\vh=(x_1,x_2,...,x_{d_\text{in}})$. Then $\gamma_u(\vh), \ u=1,2,...$ is an orthogonal basis for any Riemann integrable function on $[a,b]^{d_\text{in}}$. Thus 
\begin{equation}
    \{\gamma_{u_1}(\cdot)\cdots\gamma_{u_k}(\cdot)\},\ u_i=1,2,...,i=1,\ldots,k
\end{equation}
is an orthogonal basis for any Riemann integrable function on $[a,b]^{k\times d_\text{in}}$. To remove repeated function bases, we can define $1\leq u_1\leq u_2\leq \cdots\leq u_k$, then the set of the products $\{ \gamma_{u_1}(\cdot)\cdots\gamma_{u_k}(\cdot),$ for $1\leq u_1\leq u_2 \leq \cdots \leq u_k \} $ is an orthogonal basis for any Riemann integrable function on $[a,b]^{k\times d_\text{in}}$.

By \Cref{thm:2ary}, we can calculate the corresponding generalized Fourier coefficients. Since we combine all the repeated bases in one expression, we introduce the constants $c_{u_1,\ldots,u_k}$ here to handle it. The constants $c_{u_1,\ldots,u_k}$ is related to the repeated elements in $\{u_1,\ldots,u_k\}$. Although we do not give the explicit expression of $c_{u_1,\ldots,u_k}$, it is clear there must exist such constants to satisfy the equations.
\end{proof}

\Cref{thm:main} is a direct use of \Cref{lem:basis} as shown in the paper. We give the formal proof here.

\thmmain*

\begin{proof}
The mild assumptions stated here corresponds to the assumption in \Cref{lem:basis}, which means each component $\vec{f}_r^{(k)}$ of $\vec{f}^{(k)}$ in \Cref{def:kary} is Riemann integrable on the domain $[a,b]^{k\times d_\text{in}}$ ($a,b\in \mathbb{R}$).

We use the same definition for $\{\gamma_u(\cdot)\}_{u=1}^\infty$ and $\alpha^{(r)}_{u_1,\ldots,u_k}$ as in \Cref{lem:basis}. Plugging the representation from \Cref{lem:basis} into \Cref{eq:equaapp}, it follows that the series expansion $\dbar{f}_r^{(k)}(\vh)$, the $r$-th component of the permutation-invariant function $\dbar{f}$ takes the form: 
\begin{equation}
\begin{split}
    \dbar{f}_r^{(k)}(\vh) =& \sum_{i_1,i_2,...,i_k\in \{1,...,n_\vh\}}\vec{f}_r^{(k)}(\vh_{i_1},\vh_{i_2},...,\vh_{i_k})\\
    =&\sum_{i_1,i_2,...,i_k\in \{1,...,n_\vh\}} \lim_{M\rightarrow \infty}\sum_{u_1=1}^M  \cdots \sum_{u_k=u_{k-1}}^M \\ &{\alpha}^{(r)}_{u_1,u_2,...,u_k} \gamma_{u_1}(\vh_{i_1}) \cdots\gamma_{u_k}(\vh_{i_k})\\
    =&\lim_{M\rightarrow \infty}\sum_{u_1=1}^M  \cdots \sum_{u_k=u_{k-1}}^M {\alpha}^{(r)}_{u_1,u_2,...,u_k}\\ &\left(\sum_{i=1}^{n_\vh}\gamma_{u_1}(\vh_i)\right) \cdots \left(\sum_{i=1}^{n_\vh}\gamma_{u_k}(\vh_i)\right).
\end{split}
\end{equation}

By defining $\boldsymbol{\alpha}_{u_1,u_2,...,u_k}=[\alpha^{(1)}_{u_1,u_2,...,u_k},\cdots,\alpha^{(d_\text{rep})}_{u_1,u_2,...,u_k}]^T$, and  $\boldsymbol{\vec{\gamma}}_{u_i}(\cdot)=[\gamma_{u_i}^{(1)}(\cdot),\cdots,\gamma_{u_i}^{(d_\text{rep})}(\cdot)]^T$, we can write out a series-expansion for the entire vector-valued function $\dbar{f}^{(k)}$: 
\begin{equation}
\label{eq:hadamardapp}
\begin{split}
 \dbar{f}^{(k)}(\vh) =& \lim_{M\rightarrow \infty}\sum_{u_1=1}^M  \cdots  \sum_{u_k=u_{k-1}}^M \boldsymbol{\alpha}_{u_1,u_2,...,u_k}\odot \\&\left(\sum_{i=1}^{n_\vh}\boldsymbol{\vec{\gamma}}_{u_1}(\vh_i)\right)\odot \cdots \odot \left(\sum_{i=1}^{n_\vh}\boldsymbol{\vec{\gamma}}_{u_k}(\vh_i)\right).
 \end{split}
\end{equation}
As before, the equations above converge in mean.

If $\phi_i$ are universal approximators of $\boldsymbol{\vec{\gamma}}_{i}$ for $i=1,\ldots,M$, and $\boldsymbol{\alpha}_{u_1,u_2,...,u_k}$ are learnable parameters, then $\dbar{f}_{M,k}(\cdot)$ in \Cref{eq:kary} is a universal approximator of $\sum_{u_1=1}^M  \cdots  \sum_{u_k=u_{k-1}}^M \boldsymbol{\alpha}_{u_1,u_2,...,u_k}\odot \left(\sum_{i=1}^{n_\vh}\boldsymbol{\vec{\gamma}}_{u_1}(\vh_i)\right)\odot \cdots \odot \left(\sum_{i=1}^{n_\vh}\boldsymbol{\vec{\gamma}}_{u_k}(\vh_i)\right)$.

Thus the limit $\lim_{M \to \infty} \dbar{f}_{M,k}(\cdot)$, with $\dbar{f}_{M,k}(\cdot)$ as in \Cref{eq:kary}, converges {\em in mean} to \Cref{eq:equaapp} for any $\dbar{f}^{(k)}(\cdot)$.
\end{proof}

We now restate \Cref{thm:k_expr} and show the proof. Note that \Cref{thm:k_expr} is different from Theorem 2.1 in \citet{murphy2019janossy} because we allow repeated elements in $\{u_1,\ldots,u_k\}$ as shown in \Cref{eq:equaapp} while they focus on all possible permutations of the input sequence.

\thmexpressive*

\begin{proof}
($\mathcal{F}_{k-1}\subset \mathcal{F}_{k}$): Consider any element $\dbar{f}^{(k-1)}\in \gF$ and write $\vec{f}^{(k-1)}$ for its associated permutation sensitive function. For any sequence $\vh$, define $\downarrow_k(\vh)$ as its projection to a length $k$ sequence; in particular, if $n_\vh>k$, keep the first $k$ elements.

Then we can have

$\vec{f}^{(k-1)}(\vh_{i_1},...,\vh_{i_{(k-1)}})=\vec{f}^{(k-1)}(\downarrow_{k-1}(\vh_{i_1},...,\vh_{i_{(k-1)}},\vh_{i_k})):=n_\vh\vec{f}^{(k)}_{+}(\vh_{i_1},...,\vh_{i_{(k-1)}},\vh_{i_k})$, $\forall i_1,i_2,...,i_k\in \{1,...,n_\vh\}$, where we define the function $\vec{f}_{+}^{(k)}$ over $k$ elements but only looks at its first $k-1$ elements. 
\begin{equation}
\begin{split}
    \dbar{f}^{(k-1)}(\vh)=&\sum_{i_1,...,i_{(k-1)}\in \{1,...,n_\vh\}}\vec{f}^{(k-1)}(\vh_{i_1},...,\vh_{i_{(k-1)}})\\ =&\frac{1}{n_\vh}\sum_{i_1,...,i_{(k-1)},i_k\in \{1,...,n_\vh\}}\vec{f}^{(k-1)}\\&(\downarrow_{k-1}(\vh_{i_1},...,\vh_{i_{(k-1)}},\vh_{i_k}))\\
    =&\sum_{i_1,...,i_k\in \{1,...,n_\vh\}}\vec{f}_{+}^{(k)}(\vh_{i_1},...,\vh_{i_k})=\dbar{f}^{(k)}(\vh)
\end{split}
\end{equation}
where $\dbar{f}^{(k)}\in \gF_k$.

($\mathcal{F}_{k}\not\subset \mathcal{F}_{k-1}$) We need to find $\dbar{f}^{(k)}\in \gF_k$ such that $\dbar{f}^{(k-1)}\neq \dbar{f}^{(k)}$ for all $\dbar{f}^{(k-1)}\in \gF_{k-1}$. Let $\dbar{f}^{(k)}$ and $\dbar{f}^{(k-1)}$ be associated with $\vec{f}^{(k)}$ and $\vec{f}^{(k-1)}$, respectively.

Without loss of generality, consider $n_\vh=k$. Let $\vec{f}^{(k)}(\vh)=\prod_{i=1}^{n_\vh}\vh_i$, assuming all the elements $\vh_i$ are just scalars. So the resulting $\dbar{f}^{(k)}(\vh)$ must be continuous and differentiable with respect to $\vh_i, i=1,...,k$. Thus $\frac{\partial \dbar{f}^{(k)}(\vh)}{\partial \vh_1\partial \vh_2\cdots\partial \vh_k}= k!$. 

Assume there exits a function $\dbar{f}^{(k-1)}(\vh)\in \gF_{k-1}$, such that $\dbar{f}^{(k-1)}= \dbar{f}^{(k)}$. Then for any $\vh$, the derivative with respect to the elements should also be the same. However, since $\vec{f}^{(k-1)}$ takes at most $k-1$ distinct variables, $\frac{\partial \dbar{f}^{(k-1)}(\vh)}{\partial \vh_1\partial \vh_2\cdots\partial \vh_k}= 0$, which contradicts our assumption. So there is not such a function in $\gF_{k-1}$. We conclude our proof that $\mathcal{F}_{k}\not\subset \mathcal{F}_{k-1}$.
\end{proof}

Finally we restate \Cref{prop:univ_apprx} and show the proof.

\univ*

\begin{proof}
If we set all the coefficients  $\boldsymbol{\alpha}_{u_1,u_2,...,u_k}$ to be $\vec{0}\in \sR^{d_\text{rep}}$ except $\boldsymbol{\alpha}_{1,2,...,k}=\vec{1}\in \sR^{d_\text{rep}}$, and let $\phi_2,\cdots,\phi_k$ all output $\vec{1}\in \sR^{d_\text{rep}}$ (can be satisfied by the universal approximation ability of MLP), then the \Twister has the exact same structure as DeepSets~\citep{zaheer2017deep}, which means DeepSets is included in the structure of \Twister for any choices of $k$ and $M$. The remainder of the proof follows from the proof discussed in the Appendix of \citet{zaheer2017deep}. Hence \Twister is a universal approximator of continuous permutation invariant functions.
\end{proof}

\section{Implementation Details for synthetic tasks}

We discussed most of our implementation details in the paper. We compare \Twister's performance against widely used permutation-invariant representations on a variety of tasks for which we know the task's high-order dependencies: \DeepSets~\citep{zaheer2017deep}, $2$-ary Janossy Pooling~\citep{murphy2019janossy}, Full Janossy Pooling using GRUs with attention mechanisms (JP Full)~\citep{meng2019hats} and Set Transformer without inducing points~\citep{pmlr-v97-lee19d}. We extended the code from \citet{murphy2019janossy} for most of our tasks. We chose to use $\texttt{tanh}$ activation over $\texttt{relu}$ activation in most of our tasks because of better performance. For optimization, we searched over Adam and SGD with momentum (equal to $0.9$), and $\{1\times 10^{-4}, 5\times 10^{-4}, 1\times 10^{-3}, 5\times 10^{-3}\}$ for learning rate to achieve better performance. For the loss function, we always used the L1 loss in our experiments. In the {\em maxmin} task, we used a learning rate scheduler which decrease the learning rate by a factor of $0.9$ if the validation accuracy has stopped improving in the past $500$ epochs. Training was performed on NVIDIA GeForce RTX 2080 Ti GPUs.

When comparing with all the other baselines, we made sure the number of parameters was comparable (see \Cref{table:imgparameters,table:maxminparameters}). Since we use different neural network structures for the {\em maxmin} task with various sequence lengths over randomly encoded inputs, we report the number of the parameters in a separate \Cref{table:maxminparameters}. As we can see, \Twister always has the smallest number of parameters. Note that although theoretically, \Twister will have much more parameters if $M$ and $k$ are large. However, by setting $d_\text{rep}=d_\text{DSrep}/M$ as shown in \Cref{fig:illustration}(a), the number of trainable parameters for \Twister is much less than the corresponding DeepSets model when $k=2$. 

\begin{table}[t!!]
\vspace{-10pt}
\centering
\caption{Number of trainable parameters for different models using image (all tasks) and randomly encoded inputs (variance and range tasks). MC indicates a Monte Carlo approximation.
}
\vspace{5pt}
\label{table:imgparameters}
\resizebox{0.48\textwidth}{!}{
\begin{tabular}{llrrrrrrr}
\multicolumn{1}{c}{Cost} & 
\multicolumn{1}{c}{Model} & 
\multicolumn{1}{c}{Image inputs} &
\multicolumn{1}{c}{Encoded integers} &
\\
\toprule
\multirow{2}{*}{$O(n_\vh)$} & {\bf \Twister ($M\!=\!\!k\!=\!2$)}  & 272011 & 6031\\
& DeepSets  & 277861 &  9481\\
\hline
\multirow{2}{*}{$O(n_\vh^2)$} & JP 2-ary  & 513061 & 9481\\
 &Set Transformer & 385281 & 40833 \\
\hline
MC & JP Full (GRU)  &  433622 &  102122\\
\bottomrule
\end{tabular}
}
\vspace{-2pt}
\end{table}

\begin{table}[t!!]
\vspace{-10pt}
\centering
\caption{Number of trainable parameters for different models using randomly encoded inputs in the maxmin task with various sequence length. MC indicates a Monte Carlo approximation.
}
\vspace{5pt}
\label{table:maxminparameters}
\resizebox{0.45\textwidth}{!}{
\begin{tabular}{llrrrrrrr}
\multicolumn{1}{c}{Cost} & 
\multicolumn{1}{c}{Model} & 
\multicolumn{1}{c}{$n_\vh=10$} &
\multicolumn{1}{c}{$n_\vh=20$} &
\\
\toprule
\multirow{3}{*}{$O(n_\vh)$} & {\bf \Twister ($M\!=\!\!k\!=\!2$)}  & 12221 & 25391\\
 & {\bf \Twister ($M\!=\!3, k\!=\!2$)}  & 11171 & 22091\\
& DeepSets  & 15371 &  35291\\
\hline
\multirow{2}{*}{$O(n_\vh^2)$} & JP 2-ary  & 15371 & 35291\\
 &Set Transformer & 40833 & 40833 \\
\hline
MC & JP Full (GRU)  &  100572 &  100572\\
\bottomrule
\end{tabular}
}
\vspace{-2pt}
\end{table}

For the implementation details of \Twister, we use a mask matrix to update only sub-elements of the weight matrix as shown in \Cref{fig:illustration}(a). For $k=2$, instead of restricting $u_1\leq u_2\leq M$ as discussed in \Cref{eq:kary}, we allow repeated products by setting $u_1\leq M, u_2\leq M$ for computation simplicity. To be more specific, $\boldsymbol{\alpha}_{u_1,u_2}$ and $\boldsymbol{\alpha}_{u_2,u_1}$ will be considered as different learnable coefficients when $u_1\neq u_2$. The number of coefficients will increase from ${k+M-1\choose M-1} d_\text{rep}$ to $M^k d_\text{rep}$. However, in the case of $k=2$ with small $M$ and setting $d_\text{rep}=d_\text{DSrep}/M$, it will not cause problems as shown in \Cref{table:imgparameters,table:maxminparameters}. For Full Janossy Pooling, we use bidirectional GRU with attention as stated in \citet{meng2019hats}, and use $1$ randomly chosen permutations at test time for Monte Carlo estimation.

\section{Implementation Details for node classification tasks}
All neural network approaches, including the models proposed in this paper, are implemented in PyTorch~\citep{NEURIPS2019_9015} and Pytorch Geometric~\citep{Fey/Lenssen/2019}.

Our GCN~\citep{Kipf2016} and GAT~\citep{velickovic2018graph} implementations are based on their Pytorch Geometric implementations.
In table \Cref{tabel:graph-res}, the results for DeepSets and \Twister are reported using the best performed aggregation ({\em sum} or {\em mean}) in validation. For all the other models, we use the code in github and make sure we follow the model architecture.

The number of hidden layers was chosen between \{$2,3$\}. Specifically, the $\phi$ and $\rho$ neural network in DeepSets and \Twister can both have \{$2,3$\} hidden layers. The number of neurons in the hidden layers was chosen between \{$16,32$\} for GAT (with $8$ attention heads) and \{$128,256,512$\} for all other models. For all models we used Dropout with probability $0.5$. We optimized all models using Adam with learning rate chosen from \{$1\!\times\! 10^{-3},5\!\times\! 10^{-3},1\!\times\! 10^{-2}$\} and strength of weight decay was set as $5\!\times\! 10^{-4}$. We trained all models in a full-batch end-to-end manner except C\&S and SOTA reported from \citet{huang2020combining}. Our results show the test accuracy from the model achieving best validation accuracy. Early stopping with patience $200$ was also used.

For \Twister, since we have various sized neighborhood in the graph data, the implementation is slightly different than the synthetic tasks. We will fast pass all node features into a $\phi$ neural network. And then, we will create a mask matrix which encodes the neighborhood nodes for each node (edge relationship) as a sparse adjacency matrix, so only the values of the edge indexes are $1$. Then using this sparse adjacency matrix, we can do a matrix multiplication with the obtained $\phi$ representation for all the nodes, we get a summed (or mean) representation from the neighborhood nodes for all nodes. Finally, we concatenate the aggregated feature in the neighborhood and the original node representation, and feed into another $\rho$ neural network.

\section{More experimental results}

\subsection{The ability of \Twister to learn 2-ary dependencies}

\begin{table*}[t!!]
\vspace{-2pt}
\centering
\caption{Test accuracy and per-epoch training time in the maxmin task under different sequence lengths. Standard deviations computed over 5 runs are shown in parentheses. Best results (with significance determined by a two-sample $t$-test at $p=0.05$) are in bold. }
\vspace{5pt}
\label{tabel:2}
\resizebox{0.8\textwidth}{!}{
\begin{tabular}{llrrrr}
& & 
\multicolumn{2}{c}{$n_\vh$=10} & \multicolumn{2}{c}{$n_\vh$=20} \\
\cmidrule(lr){3-4}
\cmidrule(lr){5-6}
\multicolumn{1}{c}{Cost} & 
\multicolumn{1}{c}{Model} & 
\multicolumn{1}{c}{acc.$\uparrow$} &
\multicolumn{1}{c}{time (s/epoch)} &
\multicolumn{1}{c}{acc.$\uparrow$} &
\multicolumn{1}{c}{time (s/epoch)} \\
\toprule
\multirow{3}{*}{$O(n_\vh)$} & {\bf \Twister ($M\!=\!\!k\!=\!2$)}  & 0.705(0.011) & 1.746(0.017) & 0.616(0.027) & 1.768(0.015)\\
& {\bf \Twister ($M\!=3,\ k\!=\!2$)}  &  0.702(0.007) & 1.751(0.010) & 0.632(0.012) & 1.748(0.015)\\
& DeepSets  & 0.615(0.009) & 1.246(0.003) & 0.538(0.012) & 1.251(0.005)\\
\hline
\multirow{2}{*}{$O(n_\vh^2)$} &
JP 2-ary & 0.612(0.015)& 1.248(0.006) & 0.438(0.014) & 1.248(0.009)\\
& Set Transformer &  0.587(0.017) & 6.636(0.007)& 0.392(0.012) & 6.602(0.034)\\
\hline
\multirow{1}{*}{MC}  & 
JP Full (GRU) & {\bf 0.784(0.031)}& 4.356(0.032) & {\bf 0.868(0.018)} & 4.829(0.040)\\
\bottomrule
\end{tabular}
}
\vspace{-10pt}
\end{table*}

Following \citet{murphy2019janossy}, we explore setting the upper layer $\rho$ to be a linear layer: a feed-forward layer with identity activation and output a scalar. 
This will help clarify empirically that \Twister is more expressive than DeepSets~\citep{zaheer2017deep} without the help of $\rho$. 
Comparing with $2$-ary Janossy Pooling~\citep{murphy2019janossy}, we can further show its ability to capture $2$-ary dependencies with $k=2$.

The structure of the neural network and optimization routine remains the same for all different models in different tasks as shown in \Cref{table:extra}, \Cref{table:1,tabel:2}, except the change of $\rho$. We run all the models for $1,000$ epochs. We omit the comparison with Set Transformer and Full Janossy Pooling in this case since our main purpose here is to show the ability of \Twister to capture higher order dependencies and being more expressive than DeepSets without the help of $\rho$. 

We report the results for the image inputs in \Cref{table:imglinear}. From the table, we can see that \Twister is always the best method for different tasks. 
In the $2$-ary {\em variance task}, \Twister achieves similar results with or without a nonlinear $\rho$ compared with \Cref{table:extra}, demonstrating its ability to capture $2$-ary dependencies. 
In the $n_\vh$-ary {\em range} and {\em maxmin} tasks, while \Twister can not achieve comparable results to applying a nonlinear $\rho$, it does show significant improvements over $2$-ary Janossy Pooling and DeepSets.

\begin{table}[t!!]
\vspace{-2pt}
\centering
\caption{ Test MAE in the variance task and test accuracy in the range \& maxmin task, using {image inputs} with linear $\rho$. Per-epoch training times are included for each task. Standard deviations computed over 5 random initializations are shown in parentheses. Best results (with statistical significance determined by a two-sample $t$-test at $p=0.05$) are in bold.}
\vspace{5pt}
\label{table:imglinear}
\resizebox{0.48\textwidth}{!}{
\begin{tabular}{llrrrrrrr}
&&
\multicolumn{2}{c}{variance} & \multicolumn{2}{c}{range} & \multicolumn{2}{c}{maxmin}  \\
\cmidrule(lr){3-4}
\cmidrule(lr){5-6}
\cmidrule(lr){7-8}
\multicolumn{1}{c}{Model} & \multicolumn{1}{c}{$\rho$} &
\multicolumn{1}{c}{MAE $\downarrow$} &
\multicolumn{1}{c}{time (s/epoch)} &
\multicolumn{1}{c}{acc.$\uparrow$} &
\multicolumn{1}{c}{time (s/epoch)} &
\multicolumn{1}{c}{acc.$\uparrow$} &
\multicolumn{1}{c}{time (s/epoch)} \\

\toprule
Deep Sets & \multirow{3}{*}{Linear} & 1.582(0.017) & 1.535(0.014) & 0.350(0.005)& 1.561(0.009) & 0.345(0.002) & 1.540(0.019) \\
JP 2-ary &  & 0.974(0.032) & 1.586(0.018) & 0.475(0.009) & 1.557(0.009) & 0.354(0.007) & 1.593(0.025)\\
{\bf \Twister ($M=k=2$)} & & {\bf 0.205(0.009)} & 2.085(0.007) & {\bf 0.597(0.007)} & 2.120(0.020) &  {\bf 0.389(0.004)} & 2.124(0.033) \\
\bottomrule
\end{tabular}
}
\vspace{-2pt}
\end{table}

In \Cref{table:textlinear}, we can see the results over randomly encoded inputs. Same as the previous case, \Twister outperforms in each task especially for {\em variance} task. Note that $2$-ary Janossy Pooling is inherently able to capture $2$-ary dependencies, although \Twister does not perform well on the {\em range} and {\em maxmin} tasks, being comparable with $2$-ary Janoosy Pooling also shows its ability to capture $2$-ary dependencies. 

\begin{table}[t!!]
\vspace{-2pt}
\centering
\caption{Test accuracy and per-epoch training times in range task and test MAE in the variance task over randomly encoded inputs. Standard deviations computed over 5 random initialization runs are shown in parentheses. Best results (with statistical significance determined by a two-sample $t$-test at $p=0.05$) are in bold. }
\vspace{5pt}
\label{table:textlinear}
\resizebox{0.48\textwidth}{!}{
\begin{tabular}{lrrrrrrrrr}
&&
\multicolumn{2}{c}{variance} & \multicolumn{2}{c}{range} & \multicolumn{2}{c}{maxmin} \\
\cmidrule(lr){3-4}
\cmidrule(lr){5-6}
\cmidrule(lr){7-8}
\multicolumn{1}{c}{Model} & \multicolumn{1}{c}{$\rho$} &
\multicolumn{1}{c}{MAE $\downarrow$} &
\multicolumn{1}{c}{time (s/epoch)} &
\multicolumn{1}{c}{acc.$\uparrow$} &
\multicolumn{1}{c}{time (s/epoch)} &
\multicolumn{1}{c}{acc.$\uparrow$} &
\multicolumn{1}{c}{time (s/epoch)} \\
\toprule
 Deep Sets & \multirow{3}{*}{Linear}  & 70.3280(0.6886) & 1.1869(0.0022) & 0.0396(0.0026) &  1.1687(0.0047) & 0.070(0.003) & 1.186(0.006)\\
 JP 2-ary & & 3.5374(0.1926) & 1.1758(0.0075) & {\bf 0.0865(0.0045)} & 1.1564(0.0088) & {\bf 0.074(0.004)} & 1.166(0.005) \\
 {\bf \Twister ($M=k=2$)} & & {\bf 0.3697(0.0141)} & 1.7983(0.0070) & {\bf 0.0904(0.0047)} & 1.7816(0.0251) & {\bf 0.079(0.002)} & 1.820(0.013)\\
\bottomrule
\end{tabular}
}
\vspace{-2pt}
\end{table}

\subsection{The effect of $M$ and $d_\text{rep}$ for \Twister}

We further study the effect of $M$ for \Twister. Note that we recommend to set $d_\text{rep}=d_\text{DSrep}/M$ to prevent the number of parameters exploding as we increase $M$. 
Here we also explore how setting $d_\text{rep}$ being the same affects the performance when increasing $M$.
We do this exploration on the hardest task we have, the {\em maxmin} task over random encoded integer inputs with vocabulary size $0-99$.

Since DeepSets~\citep{zaheer2017deep} corresponds to \Twister with $M=k=1$, we also include it in the comparison. 
First we use the same neural network structure and optimization routine as discussed in the paper and extend $M$ to $M=4$. In \Cref{tabel:2}, there is a typo of $k=3$ which should be $k=2$, and we fix it in \Cref{tabel:extendm}. From \Cref{tabel:extendm}, we can see that when the sequence length is equal to $10$, increasing $M$ to $M=4$ will significantly decrease the test accuracy. One possible reason is that the neural network structure for $\phi$ is $[60,60]$ for DeepSets in this case, which means $d_\text{rep}=60/4=15$ when we set $M=4$. The small representation dimension thus restricts the capacity of the model. For sequence length equal to $20$, increasing $M$ shows improvement in performance with the current structure of $\phi$ being $[120,120]$ for DeepSets. However, the performance improvement is not significant, showing that the simplest case $M=2$ is enough to capture $2$-ary dependencies.

\begin{table}[t!!]
\vspace{-2pt}
\centering
\caption{Test accuracy and per-epoch training time in the maxmin task under different sequence lengths. Standard deviations computed over 5 runs are shown in parentheses. Best results (with significance determined by a two-sample $t$-test at $p=0.05$) are in bold. }
\vspace{5pt}
\label{tabel:extendm}
\resizebox{0.48\textwidth}{!}{
\begin{tabular}{llrrrr}
&  
\multicolumn{2}{c}{$n_\vh$=10} & \multicolumn{2}{c}{$n_\vh$=20} \\
\cmidrule(lr){2-3}
\cmidrule(lr){4-5}
\multicolumn{1}{c}{Model} & 
\multicolumn{1}{c}{acc.$\uparrow$} &
\multicolumn{1}{c}{time (s/epoch)} &
\multicolumn{1}{c}{acc.$\uparrow$} &
\multicolumn{1}{c}{time (s/epoch)} \\
\toprule
DeepSets  & 0.615(0.009) & 1.246(0.003) & 0.538(0.012) & 1.251(0.005)\\
{\bf \Twister ($M\!=\!\!k\!=\!2$)}  & {\bf 0.705(0.011)} & 1.746(0.017) & {\bf 0.616(0.027)} & 1.768(0.015)\\
{\bf \Twister ($M\!=3,\ k\!=\!2$)}  & {\bf 0.702(0.007)} & 1.751(0.010) & {\bf 0.632(0.012)} & 1.748(0.015)\\
{\bf \Twister ($M\!=4,\ k\!=\!2$)} &  0.669(0.008) & 1.759(0.013) &  {\bf 0.650(0.019)} & 1.847(0.007)\\

\bottomrule
\end{tabular}
}
\vspace{-2pt}
\end{table}

Next we set $d_\text{rep}$ to be the same for any $M$. In our case $d_\text{rep}=d_\text{DSrep}=120$. Now the first and second hidden layers of the neural network will have $[120M, 120M]$ hidden neurons. We use $\texttt{relu}$ activation in this case since it gives better validation results, while keeping all the other optimization routines the same. We report the results in \Cref{tabel:relufixmaxmin}. From the table, we can see the accuracy decreases when increasing $M$. For sequence length equal to $20$, the accuracy is significant worse than the previous cases in \Cref{tabel:extendm}. We observe overfitting problems due to the increased number of parameters, which can be solved by more training data. 

\begin{table}[t!]
\vspace{-2pt}
\centering
\caption{Test accuracy and per-epoch training time in the maxmin task under different sequence lengths when keeping $d_\text{rep}$ the same. Standard deviations computed over 5 runs are shown in parentheses. Best results (with significance determined by a two-sample $t$-test at $p=0.05$) are in bold.}
\vspace{5pt}
\label{tabel:relufixmaxmin}
\resizebox{0.48\textwidth}{!}{
\begin{tabular}{llrrrr}
& 
\multicolumn{2}{c}{$n_\vh$=10} & \multicolumn{2}{c}{$n_\vh$=20} \\
\cmidrule(lr){2-3}
\cmidrule(lr){4-5}

\multicolumn{1}{c}{Model} & 
\multicolumn{1}{c}{acc.$\uparrow$} &
\multicolumn{1}{c}{time (s/epoch)} &
\multicolumn{1}{c}{acc.$\uparrow$} &
\multicolumn{1}{c}{time (s/epoch)} \\
\toprule
{\bf \Twister ($M\!=\!\!k\!=\!2$)}  & {\bf 0.694(0.008)} & 1.708(0.033) & {\bf 0.454(0.022)} & 1.828(0.008) \\
{\bf \Twister ($M\!=3,\ k\!=\!2$)}  &  0.645(0.022) & 1.795(0.012) & {\bf 0.439(0.019)} & 1.800(0.016)\\
{\bf \Twister ($M\!=4,\ k\!=\!2$)}  &  0.617(0.012) & 1.841(0.009) & {\bf 0.437(0.026)} & 1.852(0.011) \\

\bottomrule
\end{tabular}
}
\vspace{-2pt}
\end{table}

\begin{figure}[t!]
    \centering
    \includegraphics[scale=0.2]{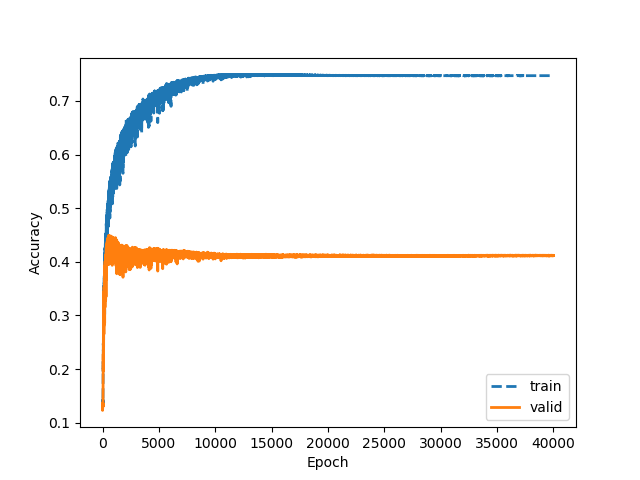}
    \includegraphics[scale=0.2]{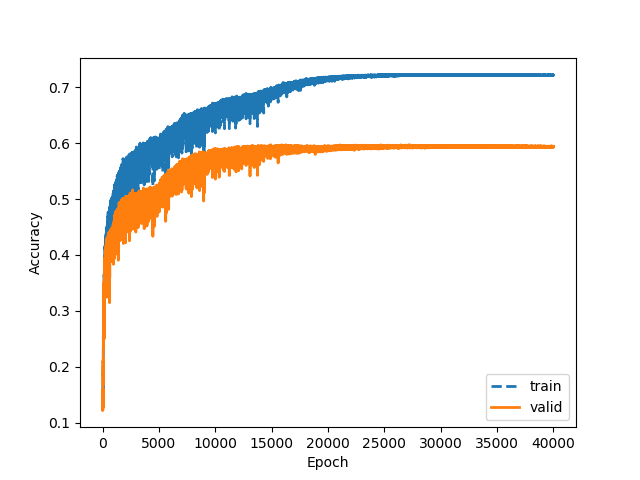}
    \includegraphics[scale=0.2]{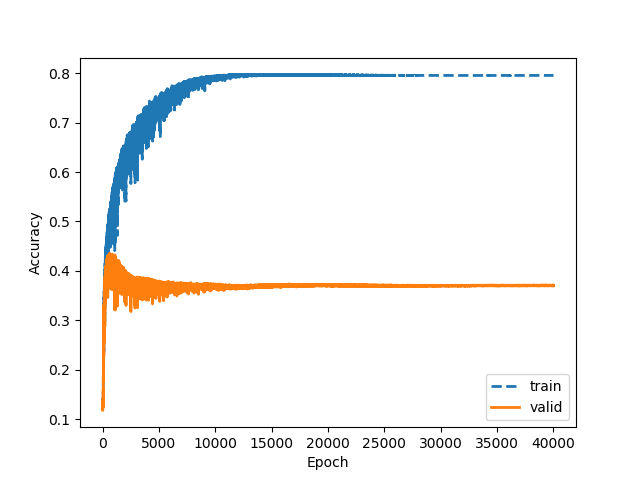}
    \includegraphics[scale=0.2]{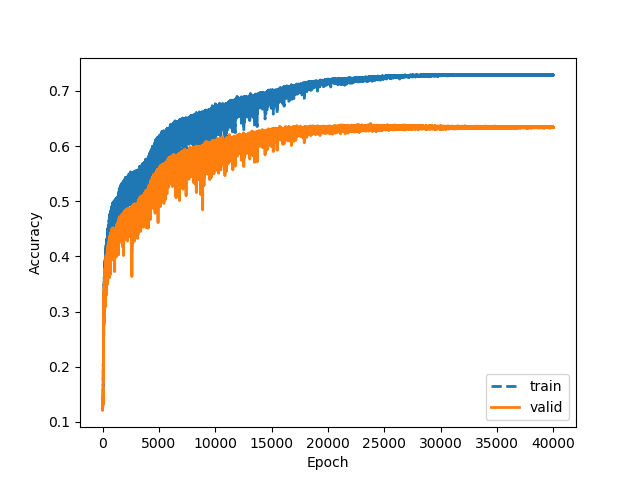}
    \includegraphics[scale=0.2]{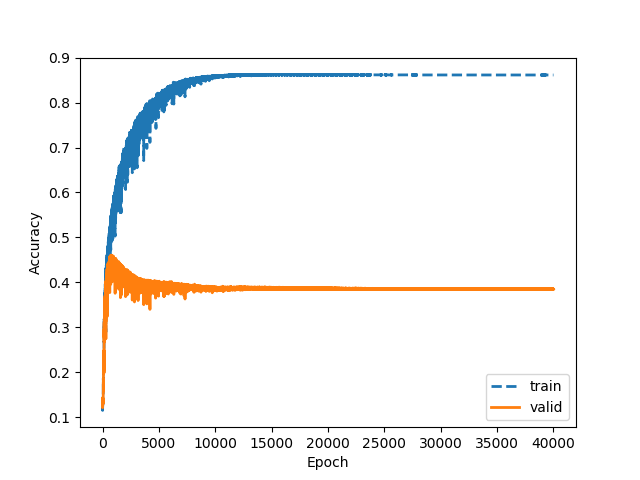}
    \includegraphics[scale=0.2]{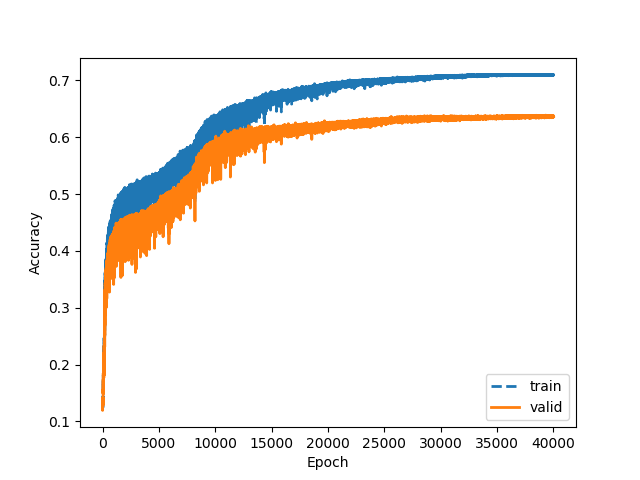}
    
    \caption{Training and validation accuracy for {\em maxmin} task over randomly encoded inputs with sequence length equal to $20$. All the plots in the first column correspond to models with fixed $d_\text{rep}$, while the plots in the second column correspond to models with decreased $d_\text{rep}$. The first row represents models with $M=2$, the second row represents models with $M=3$, and the third row represents models with $M=4$.}
    \label{fig:maxmin}
\end{figure}

To visually understand the improvement in the model capacity, we plot the training accuracy and validation accuracy for different models in \Cref{fig:maxmin}. In the left column, we can see the improvement over training accuracy with the increase of $M$, meaning the model can fit the training data better with the increase of $M$ when $d_\text{rep}$ is fixed. In the right column we do not see improvement of model capacities.

\subsection{Comparison of \texttt{tanh} and \texttt{relu} activations}

In most of our experiments, we use the $\texttt{tanh}$ activation function for DeepSets and \Twister as in \citet{zaheer2017deep,murphy2019janossy}. 
Below, we also we report results using $\texttt{relu}$ activations for DeepSets and \Twister with all the other settings the same and show the empirical advantage of using $\texttt{tanh}$ activation in \Cref{table:reluimage}, \Cref{table:relutext,table:relumaxmin}.

\begin{table}[H]
\vspace{-2pt}
\centering
\caption{Test MAE in the variance task and test accuracy in the range \& maxmin task, using {image inputs} for $\texttt{relu}$ and $\texttt{tanh}$ activations. Per-epoch training times are included for each task. Standard deviations computed over 5 random initializations are shown in parentheses. Best results (with statistical significance determined by a two-sample $t$-test at $p=0.05$) are in bold.}
\vspace{5pt}
\label{table:reluimage}
\resizebox{0.48\textwidth}{!}{
\begin{tabular}{llrrrrrrrr}
& & 
\multicolumn{2}{c}{variance} & \multicolumn{2}{c}{range} & \multicolumn{2}{c}{maxmin} \\
\cmidrule(lr){3-4}
\cmidrule(lr){5-6}
\cmidrule(lr){7-8}
\multicolumn{1}{c}{Act.} & 
\multicolumn{1}{c}{Model} & 
\multicolumn{1}{c}{MAE$\downarrow$} &
\multicolumn{1}{c}{time (s/epoch)} &
\multicolumn{1}{c}{acc.$\uparrow$} &
\multicolumn{1}{c}{time (s/epoch)} &
\multicolumn{1}{c}{acc.$\uparrow$} &
\multicolumn{1}{c}{time (s/epoch)} \\
\toprule
\multirow{2}{*}{$\texttt{relu}$} & {\bf \Twister ($M\!=\!\!k\!=\!2$)}  & 0.643(0.038) & 2.376(0.015)  & 0.760(0.025)  & 2.444(0.029) & 0.332(0.005) & 2.425(0.024)\\
& DeepSets  &  0.656(0.020) & 1.798(0.005) & 0.906(0.007) & 1.782(0.026) & 0.358(0.046)& 1.727(0.043)\\
\hline
\multirow{2}{*}{$\texttt{tanh}$} & {\bf \Twister ($M\!=\!\!k\!=\!2$)}  & {\bf 0.200(0.007)} & 2.386(0.020) & 0.931(0.002) & 2.441(0.036) & {\bf 0.931(0.003)} & 2.428(0.049) \\
& DeepSets  & 0.343(0.012) & 1.804(0.015) & {\bf 0.940(0.001)} & 1.842(0.012) &  0.912(0.003) & 1.839(0.012)   \\
\bottomrule
\end{tabular}
}
\vspace{-2pt}
\end{table}

\begin{table}[H]
\vspace{-2pt}
\centering
\caption{Test accuracy and per-epoch training times in range and maxmin tasks and test MAE in the variance task over randomly encoded inputs for $\texttt{relu}$ and $\texttt{tanh}$ activations. Standard deviations computed over 5 random initialization runs are shown in parentheses. Best results (with statistical significance determined by a two-sample $t$-test at $p=0.05$) are in bold.}
\vspace{5pt}
\label{table:relutext}
\resizebox{0.48\textwidth}{!}{
\begin{tabular}{llrrrr}
& & 
\multicolumn{2}{c}{variance} & \multicolumn{2}{c}{range} \\
\cmidrule(lr){3-4}
\cmidrule(lr){5-6}
\multicolumn{1}{c}{Act.} & 
\multicolumn{1}{c}{Model} & 
\multicolumn{1}{c}{MAE$\downarrow$} &
\multicolumn{1}{c}{time (s/epoch)} &
\multicolumn{1}{c}{acc.$\uparrow$} &
\multicolumn{1}{c}{time (s/epoch)} \\
\toprule
\multirow{2}{*}{$\texttt{relu}$} & {\bf \Twister ($M\!=\!\!k\!=\!2$)}  & {\bf 0.181(0.020)} & 2.034(0.003) & 0.808(0.032)  & 2.128(0.011)\\
& DeepSets  & 0.331(0.045)  & 1.522(0.010) & 0.861(0.013) & 1.506(0.008)\\
\hline
\multirow{2}{*}{$\texttt{tanh}$} &
{\bf \Twister ($M\!=\!\!k\!=\!2$)}  & 0.333(0.008) & 2.072(0.011) & {\bf 0.944(0.004)} & 2.105(0.018) \\
& DeepSets & 0.417(0.015) & 1.501(0.016) & {\bf 0.943(0.004)} & 1.504(0.006) \\
\bottomrule
\end{tabular}
}
\vspace{-2pt}
\end{table}

\begin{table}[H]
\vspace{-2pt}
\centering
\caption{Test accuracy and per-epoch training time in the maxmin task under different sequence lengths for $\texttt{relu}$ and $\texttt{tanh}$ activations. Standard deviations computed over 5 runs are shown in parentheses. Best results (with significance determined by a two-sample $t$-test at $p=0.05$) are in bold. }
\vspace{5pt}
\label{table:relumaxmin}
\resizebox{0.48\textwidth}{!}{
\begin{tabular}{llrrrr}
& & 
\multicolumn{2}{c}{$n_\vh$=10} & \multicolumn{2}{c}{$n_\vh$=20} \\
\cmidrule(lr){3-4}
\cmidrule(lr){5-6}
\multicolumn{1}{c}{Act.} & 
\multicolumn{1}{c}{Model} & 
\multicolumn{1}{c}{acc.$\uparrow$} &
\multicolumn{1}{c}{time (s/epoch)} &
\multicolumn{1}{c}{acc.$\uparrow$} &
\multicolumn{1}{c}{time (s/epoch)} \\
\toprule
\multirow{4}{*}{$\texttt{relu}$} & {\bf \Twister ($M\!=\!\!k\!=\!2$)}  & 0.679(0.009)& 1.770(0.013) & {\bf 0.618(0.033)} & 1.773(0.012)\\
& {\bf \Twister ($M\!=3,\ k\!=\!2$)}  & 0.648(0.007)   & 1.814(0.015) & {\bf 0.638(0.013)} & 1.678(0.012) \\
& {\bf \Twister ($M\!=4,\ k\!=\!2$)}  & 0.620(0.031)  & 1.824(0.008) & {\bf 0.616(0.019)} & 1.767(0.012)\\
& DeepSets  & 0.596(0.006)  & 1.213(0.002) & 0.517(0.016) & 1.230(0.005)\\
\hline
\multirow{4}{*}{$\texttt{tanh}$} & {\bf \Twister ($M\!=\!\!k\!=\!2$)}  & {\bf 0.705(0.011)} & 1.746(0.017) & {\bf 0.616(0.027)} & 1.768(0.015)\\
& {\bf \Twister ($M\!=3,\ k\!=\!2$)}  & {\bf 0.702(0.007)} & 1.751(0.010) & {\bf 0.632(0.012)} & 1.748(0.015)\\
& {\bf \Twister ($M\!=4,\ k\!=\!2$)} &  0.667(0.011) & 1.759(0.013) &  {\bf 0.650(0.019)} & 1.847(0.007)\\
& DeepSets  & 0.615(0.009) & 1.246(0.003) & 0.538(0.012) & 1.251(0.005)\\
\bottomrule
\end{tabular}
}
\vspace{-2pt}
\end{table}

From the tables, we can see using $\texttt{tanh}$ activation gives significantly better results in most of the tasks, especially for the image inputs, which justifies our use of $\texttt{tanh}$ activation.

\end{document}